\documentclass[10pt,twocolumn,letterpaper]{article}

\usepackage{cvpr}              % To produce the CAMERA-READY version
\newcommand{\Major}[1]{{}}
 
\newcommand{\KG}[1]{{\color{cyan}#1}} %magenta
\newcommand{\JW}[1]{{\color{orange}#1}}
\renewcommand{\JW}{\textcolor{black}}
\renewcommand{\KG}{\textcolor{black}}

\newcommand{\JWnew}{\textcolor{black}}

\newcommand{\JWCam}[1]{{\color{black}#1}}

\usepackage{amsmath}

\usepackage{array}
\newcommand{\PreserveBackslash}[1]{\let\temp=\\#1\let\\=\temp}
\newcolumntype{C}[1]{>{\PreserveBackslash\centering}p{#1}}
\newcolumntype{L}[1]{>{\PreserveBackslash\raggedright}p{#1}}
\usepackage{booktabs}
\usepackage{pifont}
\newcommand{\cmark}{\ding{51}}
\newcommand{\xmark}{\ding{55}}
\usepackage{multirow}

\definecolor{cvprblue}{rgb}{0.21,0.49,0.74}
\usepackage[pagebackref,breaklinks,colorlinks,allcolors=cvprblue]{hyperref}

\def\paperID{XXXX} % *** Enter the Paper ID here
\def\confName{CVPR}
\def\confYear{2026}

\title{SkillSight: Efficient First-Person Skill Assessment with Gaze}
\author{Chi Hsuan Wu, Kumar Ashutosh, Kristen Grauman\\
Univeristy of Texas at Austin\\
}

\begin{document}
\maketitle
\begin{abstract}

\vspace{-0.4cm}

Egocentric perception on smart glasses could transform how we learn new skills in the physical world, but automatic % However,
skill assessment
%\cc{---an essential capability for adaptive guidance, progress tracking, and timely intervention---} 
remains a fundamental technical challenge.
%Smart glasses are rapidly emerging as a platform for %skill learning and providing real-world assistance, 
%where skill assessment forms the basis for adaptive guidance, progress tracking, and timely intervention. 
%We reveal that gaze, inherently available on many smart glasses, provides a powerful cue for assessing expertise. 
We introduce SkillSight for power-efficient skill assessment from first-person data.  
Central to our approach is the hypothesis that skill level is evident not only in how a person performs an activity (video), but also in how they direct their attention when doing so (gaze).  
Our two-stage framework %for power-efficient skill assessment that 
first learns to jointly model gaze and egocentric video when predicting  skill level, % exhibited in an input sequence, 
then distills a gaze-only student model. % from the teacher. 
At inference, the student model requires only gaze input, %---lightweight and readily  available on most smart glasses---
drastically reducing power consumption by eliminating continuous video processing. Experiments on three datasets spanning cooking, music, and sports establish, for the first time, the valuable role of gaze in skill understanding across diverse real-world settings. Our SkillSight teacher model achieves state-of-the-art performance, while our gaze-only student  variant maintains high accuracy using 73$\times$ less power than competing methods.  These results pave the way for in-the-wild AI-supported skill learning.
%ranks second in power consuming baselines at over 
%70× lower power, pushing the boundaries of power efficiency and accuracy.
%\KGnote{the power part comes out strongly in abstract; I think we also want to emphasize how we establish the role of gaze for skill estimation in variety of in-the-wild, real-world settings for the first time. and push state of art for gaze skill estimation from video (provided our V+G results are top).}
\footnotetext[1]{Project page: \href{https://vision.cs.utexas.edu/projects/skillsight/}{https://vision.cs.utexas.edu/projects/skillsight/}}

% \footnotetext[1]{Equal contribution.} \footnotetext{Project page: \href{https://vision.cs.utexas.edu/projects/stitch-a-demo/}{https://vision.cs.utexas.edu/projects/stitch-a-demo/}}
\end{abstract}
    
\vspace*{-0.13in}
\section{Introduction}
\label{sec:intro}

\begin{figure}[t] % [t] = top of column, can also use [h] or [b]
    \centering
    \includegraphics[width=\linewidth]{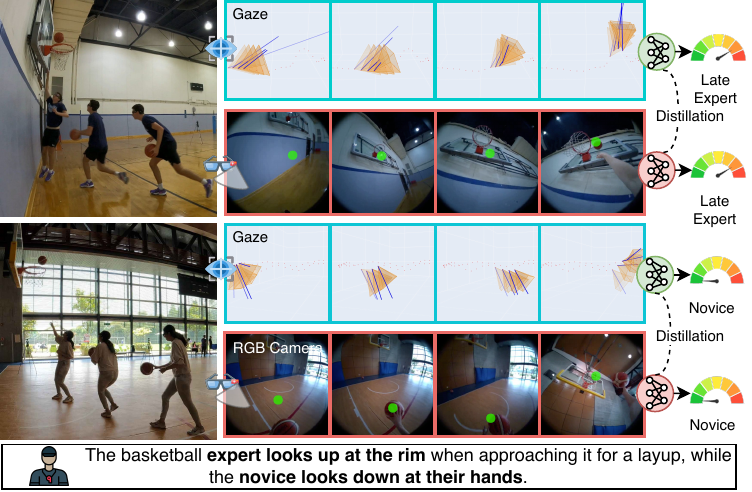}
    % Placeholder box
    % Caption
    \caption{\textbf{Skill assessment with gaze.}  
    Experts and novices exhibit distinct %gaze 
    attention behaviors, influencing both how they move their head and eyes and what they see,     
    as illustrated here with clips from an expert (top) and novice (bottom) basketball layup from~\cite{egoexo4d}. %Egocentric visual with gaze signals highlights where a person is looking. Our 
    The proposed method explores the %correlation between 
    associations between gaze, action, and expertise %. Furthermore, we find that gaze patterns provide rich information about an individual's skill level, enabling 
    to achieve accurate and power-efficient skill assessment, using either ego-video and gaze, or gaze alone.  \JW{The blue ray indicates gaze direction and depth, while shading shows \KG{camera} %glass
    motion over past frames.}  Note: leftmost third-person timelapses \KG{and commentary text} are for illustration only.
    % \KGnote{fix the gauge figs to be low for novice and high for expert} \JWnote{I fixed the novice figs, is it clear now?}  \KGnote{I meant that all the dials point to the same red region on right hand side.}
    %\KGnote{throughout I think we should claim both we're doing something good by augmenting video with gaze \emph{and} the low power part.  don't understate the gaze+video findings. brighten the RGB images?}
    }
    \label{fig:intro}
    \vspace{-0.5cm}
\end{figure}

%Smart glasses are becoming increasingly popular. 
Egocentric perception is poised to transform AI assistants on smart glasses which, by seeing through the eyes of a user, could provide in-the-moment contextually relevant information and recommendations. Of particular interest are assistants to support learning new skills 
%An important application of smart glasses is to learn new skills 
in various domains such as exercise, sports, cooking, and music% with AR guidance
~\cite{vid2coach,prosandcons,evostruggle,whentosay,fineparser,expertaf,egoexo4d,egoexolearn, holoassist}. 
% \KGnote{add holoassist ref if not here yet, and double check we're adequately citing Angela Yao's recent work in this paper (can be in Related).}
\emph{Skill assessment}---the task of quantifying the degree of skill exhibited in a given execution---%
%In this context, \emph{skill assessment} 
plays a crucial role: 
it would enable timely support~\cite{whentosay}, tracking of personal progress~\cite{ProgressAssessment}, and identifying areas for improvement~\cite{exact}. Across these capabilities and more, skill assessment has the potential to personalize learning and enhance user performance in real-world tasks.  Meanwhile, the portability of wearable glasses  opens up seamless in-the-wild capture even for dynamic physical activities that go well beyond lab environments---e.g., the soccer pitch, the dance floor, or basketball court.

%As wearable and portable devices, smart glasses naturally enable applications that seamlessly extend into in-the-wild settings. 

%\KGnote{should we also work in how glasses being a wearable/portable/always-on device means such systems could allow seamlessly venturing out in the wild where people are performing these skills?  I'm thinking to foreshadow the contrast we make later with exo, where we say there are complex device setups.}

%\KGnote{shall we move prior work up here?}
However, prior research on skill assessment
primarily relies on third-person visual perspectives of a subject's body poses \cite{logo, expertaf, multitask_assessment, videodiff, BASKET_CVPR25}, assuming prior setup of camera(s) in each target environment. %, and failing to capture %These approaches often require complex setups, such as wearing sensors or setting up cameras during exercise, and fail to capture 
%the cognitive decision-making processes critical to skill execution. 
Only limited work considers skill assessment from an egocentric perspective~\cite{egoexolearn, egoexo4d, prosandcons, AmIABaller}, and there the low visibility of the camera-wearer's full body remains a critical challenge outside of table-top settings.
%, and user attention is considered only in stationary table-top cooking and lab tasks~\cite{egoexolearn}.  
Furthermore, the high power consumption of continuous video recording is an obstacle for %all prior
vision-based methods---at odds with application needs for real-time, interactive skill learning.

Among the sensing modalities on smart glasses, we hypothesize that \emph{gaze} is uniquely informative for assessing skill.  Gaze complements vision: 
%On the one hand, an egocentric camera shows what is in view but misses some of the body cues about skill; on the other hand,  
%Gaze complements vision. 
together, they reveal not only what the user is attending to, but also their intention \cite{EyeBeholder,GIMO}. This synergy exposes fine-grained execution details that cameras alone cannot capture. 
In cognitive science, it is well known that %in decision-making, 
people often fixate on objects they intend to manipulate or evaluate~\cite{psy_decisionmaking}, while in domains as broad as
%Across domains such as 
sports~\cite{psy_gazesport}, surgery~\cite{psy_surgerytraining}, and music~\cite{eyepiano}, experts display distinctive gaze patterns that enable them to execute complex motor actions more skillfully.
%\KGnote{unclear - do we mean that an expert watching another performer pays attention to their gaze (like there's evidence human coaches are doing this?)?  but then the volleyball example seems to simply say that there are distinctions in expert vs. novice gaze patterns.}  
For example, %studies have shown that 
volleyball experts fixate earlier on the ball’s contact point with their arms compared to novices~\cite{psy_gazesport}, while skilled soccer players allocate more gaze to their surroundings while handling the ball~\cite{psy_soccer_intro}, and the final steady fixation of the \emph{quiet eye} is a signature not only of skilled athletes~\cite{Vickers1996} but also skilled surgeons~\cite{Vickers2011Surgery,Causer2014}, drivers~\cite{Vickers2016Driving}, and musicians~\cite{DraiZerbib2012}. 
Could incorporating gaze into AI skill assessment provide such access to the cognitive and motor processes underlying an individual's actions, allowing more accurate estimates? 
To this end, we introduce SkillSight, a two-stage multimodal learning framework for first-person data. 
First, we train a teacher model SkillSight-T that integrates egocentric video and gaze to capture skill-related features. SkillSight-T generalizes across in-the-wild scenes by modeling interactions between gaze and action, encoding object fixations and transitions from gaze-cropped images, and modeling the dynamic gaze patterns. In the second stage, we train a student model SkillSight-S that relies \emph{only} on gaze as input 
\KG{and  keeps the camera off during inference---significantly reducing power consumption, while also increasing user privacy.} To connect action, skill, and gaze, we train SkillSight-S via knowledge distillation, transferring visual information from SkillSight-T into gaze. Gaze signals encode spatial and temporal patterns of attention (e.g., fixations, saccades) that correlate closely with visual cues in egocentric video, enabling SkillSight-S to infer skill-related features without \KG{RGB} input.
%\KGnote{this para jumps to the nitty gritty too quickly, and again I think we should claim both 1) ego video+gaze for skill assessment in the wild and 2) the gaze-power idea.  like we introduce the base model which is great because... then we introduce the power reducing version which...}
% For example,EgoExo4D \cite{egoexo4d, Egoexo4D_15mW} approximates a power consumption rate of 15mW for RGB camera operation, compared to only 3.6mW  in total for eye tracking \cite{power_gaze_track, power_gaze_cam} and IMU \cite{power_IMU}.

%...Also, using visual information may further increase the user's privacy concern.

%\KGnote{if we say ``practical deployment" then the absolute power use vs. device abilities becomes relevant. do we have this?} \JWnote{We don't and I think it is hard to make this plausible given the information provided by different prototypes.}

% \KGnote{I would weave this part into the datasets/experiments summary below, somewhere after the 3 datasets piece.} To validate our gaze-based approach, we analyze individuals with varying skill levels, revealing significant differences in gaze metrics—such as gaze angles, saccade speed, and fixation point diversity—consistent with psychology research \cite{psy_esports, psy_gazesport, music_eyehand}. 

%\KGnote{think about what can we claim as far as novelty and depth of our quantitative analysis vs. the cog sci literature, facilitated by the ego-exo4d data.}

We evaluate our method on three datasets (Ego-Exo4D~\cite{egoexo4d}, Multisense Badminton~\cite{badminton}, Expert-Novice Soccer~\cite{expert_novice_soccer})  spanning cooking, music, and various sports. SkillSight-T outperforms previous video-based methods by 5\% \KG{(10\% relative)}. 
% \KGnote{how much relative?  if big we could put both; 5 sounds good but not overwhelming.} 
SkillSight-S, which relies solely on gaze, performs competitively while %ranks the second with around 
consuming \JW{$14\times$} to $73\times$ less power, and outperforming % It also demonstrates higher accuracy with lower power consumption compared to other 
existing methods aimed at efficiency~\cite{X3d,egodistill,egotrigger}. 
% \KGnote{check.} %efficient baselines. 
Beyond performance, we provide quantitative and qualitative analyses revealing when and how gaze reflects skill. Together, these results highlight gaze as a powerful cue for scalable skill assessment.

Overall, we pioneer skill assessment using gaze signals across diverse domains %encompassing both static and 
and dynamic in-the-wild scenarios involving significant subject motion across the scene (e.g., climbing a boulder or dribbling to the basket for a layup, as opposed to tabletop activities).  We are the first to explore power-efficient, privacy-preserving egocentric skill assessment, paving the way for practical deployment on resource-constrained smart glasses. Moreover, our analysis reveals how model predictions align with \KG{and even enhance}  established psychological theories, offering new quantitative, data-driven insights into complex gaze–skill relationships.

%\KGnote{somewhere in intro, should we motivate how gaze is especially helpful for ego video because much of the body won't be visible (handicaps the video), yet is a standard signal for understanding fine-grained details about execution that suggest skill level. this goes along with not understating even the case where we use V+G.} \JWnote{See paragraph 2 [Our insight on adding gaze.]}

%\KGnote{I don't think we're explicit enough about the skill assessment task definition in intro, i.e., given a sequence, predict the discrete skill level or skill rating... and should we be clear that we're estimating the inherent skill level of the person, not the skill observed in that particular execution?  we do the person level (I think) but technically we could handle both, which is more general and useful for the kind of coaching apps you describe at the top.} \JWnote{See paragraph 5 [High-level idea of our model.]}

\section{Related Work}
\label{sec:relatedwork}

%-------------------------------------------------------------------------
\textbf{Egocentric video and gaze.}  Gaze complements egocentric video by revealing attention and intention. Prior work predicts gaze from the ego view to model decision-making \cite{ListenToLookIntoFuture, PredictEgoGaze, EyeofTransformer, GazeFromTask} and leverages gaze for tasks such as action recognition \cite{EyeBeholder, gaze_action_recognition}, motion anticipation \cite{GIMO, GazeMotion, GazeActionPredict}, privacy filtering \cite{privaceye}, attended-object detection \cite{AttendedObjects, YoudoIlearn}, intention understanding \cite{gazegpt, egogaze}, error detection \cite{gazemissteps}, and learning sports play \cite{basketball_gaze}. However, all such work focuses on aligning gaze with actions rather than assessing performance quality. Skill assessment demands recognizing subtle behavioral differences, % between experts and novices, which is far more challenging than 
beyond simply identifying actions. We instead explore how discriminative gaze trajectories reveal expertise across diverse domains, establishing gaze as a reliable and scalable cue for skill.
%\JWnote{paragraph is shortened by rephrasing.}
% Gaze signals naturally complement egocentric video by revealing both the focus of attention and underlying intention. Some work predicts gaze from ego view to better understand human decision-making \cite{ListenToLookIntoFuture, PredictEgoGaze, EyeofTransformer, GazeFromTask}. Beyond prediction, gaze has been used to support diverse tasks such as action recognition \cite{EyeBeholder, gaze_action_recognition}, motion anticipation \cite{GIMO, GazeMotion, GazeActionPredict}, task-relevant object discovery \cite{YoudoIlearn}, privacy-sensitive content identification \cite{privaceye}, and attended-object detection \cite{AttendedObjects}. Gaze has also been utilized to better understand human intention and provide support from personal assistants \cite{gazegpt}. Other studies measure the deviation between predicted and actual gaze trajectories for mistake detection \cite{gazemissteps}, or jointly estimate the gaze of multiple players in sports scenarios \cite{basketball_gaze}. While these methods primarily aim to align gaze with actions or predict actions from gaze, they do not tackle the more challenging problem of assessing how well a task is performed. Skill assessment requires recognizing subtle behavioral differences between experts and novices, which is significantly more difficult than merely identifying the action itself. In contrast, we investigate how discriminative gaze trajectories reveal expertise across diverse domains, demonstrating gaze as a reliable and scalable indicator of skill.

\textbf{Relation of gaze and skills in cognitive science.} Existing psychology studies investigate the relationship between gaze patterns and everyday tasks \cite{psy_dailyroles}, decision-making \cite{psy_decisionmaking}, goal-directed behavior \cite{psy_GazeAndGoal}, task difficulty \cite{psy_GazeHotspot}, and anticipation of future procedural steps \cite{psy_LookAheadFixations}. As discussed above, cognitive science research links gaze patterns to proficiency: in medicine, %gaze has been used 
gaze helps assess 
diagnostic and surgery skills \cite{psy_medicaleducation, psy_robotsurgery, psy_surgerytraining}; in sports, expert athletes demonstrate distinct gaze strategies \cite{psy_gazesport, psy_esports}. We take inspiration from their findings.  Further, building on this foundation, our work enables large-scale, data-driven learning of gaze-skill relations in diverse in-the-wild settings, uncovering subtle patterns beyond controlled psychology studies. 
%\KGnote{could we be more explicit and dramatic here? claim our work marks the advent of large-scale, high quality quantitative analysis of in-the-wild activities w/skill, including scenarios with people moving about the scene?  we don't want to just re-discover what they have observed already; we want to find even more subtle patterns that emerge from data, and take the analysis out of the lab and into real-world settings...}

%-------------------------------------------------------------------------
\textbf{Skill assessment.} Prior work on skill assessment focuses on third-person pose analysis in fitness \cite{FitnessAQ}, skating \cite{skating}, and diving \cite{diving}. In contrast, first-person perspectives captured by wearables offer cues for real-time feedback in hand-centric tasks \cite{Piano, Surgery, egoexolearn,prosandcons,egoblind,egotextvqa} or sports \cite{AmIABaller, egoexo4d}. Recent work further incorporates text \cite{visualtextAQA, narrativeAQA, videotextAQA2, videotextAQA3}, audio \cite{AV-AQA, beatsAQA, multimodalAQA}, human skeletons \cite{skeleton, skeleton2, skeleton3}, PPG \cite{egoppg}, and IMU~\cite{expert_novice_soccer, IMU_Skill}.
%\JW{While some of them may be costly, gaze tracking is commonly implemented on smart glasses} \KGnote{well the gaze sensor could also be considered ``special"} 
%and gaze is widely recognized in psychology as an indicator of expertise. 
To our knowledge, \cite{egoexolearn} is the only vision work estimating skill \KG{with} gaze, and it is shown only on static tasks  (cooking and lab work) %surgery
where the subject remains stationary. Our approach instead extends gaze-based skill assessment to dynamic activities, shows broad applicability across settings, and introduces technical novelty to explicitly capture the gaze-action interplay. %\KGnote{add here to strengthen our claim; brief generic phrase here is ok.}
%\JWnote{There are a few medical papers about gaze and skill assessment, so I wrote \emph{computer vision work}.} \KGnote{meaning where the person stands in one spot? let's try to have a briefer, sharper contrast with~\cite{egoexolearn} and we can be clear that this is the only work we're aware of where skill is estimated with gaze.}

% Yet, only limited efforts \cite{egoexolearn} have incorporated gaze for skill assessment, and only in static domains such as cooking and surgery. 
% \KGnote{meaning where the person stands in one spot?}  Their approach focuses on encoding gaze regions, which is not suitable for sports scenarios where gaze and action cues are often not spatially aligned. \KGnote{unclear.} Our work systematically investigates how gaze patterns reveal expertise across both static and dynamic settings, broadening the applicability of gaze as a scalable cue for skill assessment.  \KGnote{let's try to have a briefer, sharper contrast with~\cite{egoexolearn} and we can be clear that this is the only work we're aware of where skill is estimated with gaze.}
%-------------------------------------------------------------------------

\begin{figure*}[t]
    \centering
    % Placeholder box (adjust height as needed)
    % \fbox{\rule{0pt}{2in} \rule{\textwidth}{0pt}}
    \includegraphics[width=\textwidth]{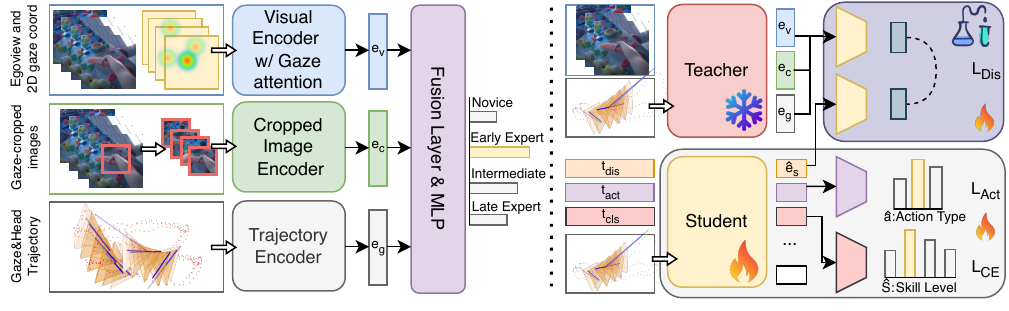}
    \vspace{-0.3in}
    \caption{\textbf{Left: Overview of SkillSight-Teacher.} We incorporate three components that encode action and gaze correlation, attended object sequence, and gaze trajectory for skill assessment. These features are fused by the fusion layer for prediction. \textbf{Right: Overview of distillation method.} \JW{SkillSight-Student learns to distill knowledge from the teacher feature $[e_v,e_c,e_g]$ using the distillation token $t_{dis}$. As guidance for evaluating skill in context, the student model performs subtask recognition with the action recognition token $t_{act}$.} 
    % \KGnote{in text it is $t_{dis}$ and $t_{act}$.} %\KGnote{label the histograms, both sides? adjust colors to avoid association between T and S rectangles.}
    %\KGnote{explain ACT better briefly here...regularization? and work in more notation: $e_v$? $e_c$?, $e_g$?}
    }
    \label{fig:model}
    \vspace{-0.4cm}
\end{figure*}

\textbf{Efficient methods for wearable devices.} Power efficiency is critical in wearable devices. Prior work addresses it through adaptive power management \cite{smartapm}, distributed computation \cite{distributedcomputation}, selectively sampling clips \cite{scsampler, flexibleframe}, and lightweight model architectures \cite{X3d, lightASDNet}.  More relevant to our work, another direction reduces reliance on power-hungry video by using lighter modalities: %. For example, previous work uses audio cues 
audio can suggest when to process video frames~\cite{listentolook, egotrigger, chat2map}, and IMU with sparse video frames is sufficient for action recognition~\cite{egodistill}. % leverage sparse frames and IMU signals for action recognition. 
Nevertheless, all these prior methods still depend on periodic visual input, requiring frequent camera toggling or low frame rate operation, which undermines both hardware simplicity and power efficiency due to startup latency \cite{projectaria_glasses_manual} and transient power spikes when switching on the camera \cite{camera_power, camera_power2}. 
%\KGnote{I feel like our contrast is bigger than this; text is understated here / this is somewhat a detail.}  
In contrast, we distill visual supervision during training but use only gaze at inference, removing the need for camera input and substantially lowering sensing and model power, as we will quantify in results.

%....Another challenge to practical smart glasses applications is the high power consumption of continuous video recording \cite{egoexo4d, egoexolearn}. While power-efficient methods like selective visual processing \cite{chat2map, egodistill, egotrigger, listentolook} reduce energy use by intermittently toggling the camera, this introduces hardware complexity such as startup latency \cite{projectaria_glasses_manual} and transient power spikes when switching on the camera \cite{camera_power, camera_power2}. \JW{These effects can ironically increase overall power consumption and cause the system to capture meaningless frames due to the delay. Also, using visual information may further increase the user's privacy concern.} 

%\KGnote{overall we should make Related about 15-20\% shorter or so.}

%-------------------------------------------------------------------------

\section{Method}
\label{sec:method}
%\KGnote{to be explicit somewhere in setup: method  can assign skill level to any temporal stretch, and it need not be constant for the same subject.  Otherwise, it's not useful for skill learning apps.}

We formally define the problem statement (Sec.~\ref{sec:problem}), then introduce our model (Sec.~\ref{sec:teacher} and~\ref{sec:distill}) and describe data and implementation details (Sec.~\ref{sec:data_detail}).

\subsection{Problem statement}\label{sec:problem}

Consider a dataset $\mathcal{E}=\{(V,G,S)\}$, where each $V_i=\{v_i^t\}_{t=1}^T$ is the egocentric video demonstration with its frames $v_i^t$, $G_i=\{g_i^t\}_{t=1}^T$ is the gaze pattern,  and $S_i$ is the skill-level of the demonstrator. Although skill is inherently complex, recent studies and datasets have introduced rigorous objective means to quantify skill~\cite{egoexo4d,badminton,expert_novice_soccer}, \KG{formalizing} this research direction.
%While skill can be subjective, we rely on consistent expert-defined proficiency labels for objective evaluation.
%subject. 

Consistent with current hardware, we suppose that the device records the glasses' rotation and translation, as well as the 3D gaze vector of each eye, from which we derive $g_i^t$, which includes the 3D fixation points, the 3D gaze direction relative to the center of two eyes, the 2D coordinate of the gaze projection on the egoview video $g_{2d}\in R^2$, the depth of the gaze, and the translation and quaternion rotation of the glass. \KG{Current devices efficiently estimate gaze with eye cameras, IR,  EOG, and/or IMU}; we detail data resources~\cite{egoexo4d, badminton, expert_novice_soccer} in Sec.~\ref{sec:data_detail} \KG{and quantify power load in Sec.~\ref{sec:expts}.} %%%\KGnote{should we add a footnote or pointer to a hardware survey to briefly summarize the typical kinds of gaze sensors, e.g., our results use devices where gaze comes from low-cost eye-facing cameras [ref]...etc.  We want to pre-empt confusion about sensor costs and where gaze comes from.} 
%\KGnote{reader will wonder where all these precise gaze readings come from; we should state at first what we assume ability to sense (no matter the sensor or platofrm that produces it), i.e., 3D gaze vector for the fixation and... (from which all the components can be derived. the we can state for clarity the specific devices/sources we'll show in expts, drawn from 3 existing datasets [refs].}
% While the third-person perspective video provides a clear human pose cue, we focus on skill assessment using modalities provided by the smart glasses, which is a setting that helps to connect with downstream tasks such as generating guidance without additional setup. \KGnote{probably we have motivated this enough already.}

The goal of this work is to classify\footnote{Similarly, one could formulate the task as regression to a real-valued score~\cite{score_skill, diving, beatsAQA}.  We target discrete classes to account for the granularity of expertise differences discernible by human judges~\cite{skillformer, skill_level, expert_novice_soccer, piano_skill_level} and to align with multiple existing annotated datasets~\cite{egoexo4d, badminton, BASKET_CVPR25}.}
%a subject's
the skill level $S_i$ using modalities from the smart glasses. We consider two setups: (1) \textbf{Video+Gaze}: we leverage both video and gaze during training and inference. Formally, we aim to learn a function $\mathcal{F}_{v}(V,G) \rightarrow S$, and call this variant of our method SkillSight-T(eacher) (2) \textbf{Gaze-only}: Continuous camera recording is power consuming and impractical for long-duration use. To reduce the reliance on camera, we use both video and gaze during training but rely only on gaze during inference. Specifically, we aim to learn $\mathcal{F}_g(G) \rightarrow S$, and call this variant of our method SkillSight-S(tudent). %While skill can be subjective, we rely on consistent expert-defined proficiency labels for objective evaluation.
%\KGnote{around here or intro, should we get in front of criticism that skill level can be subjective, etc.?} 

\begin{figure*}[t]
    \centering
    % Placeholder box (adjust height as needed)
    \includegraphics[width=\textwidth]{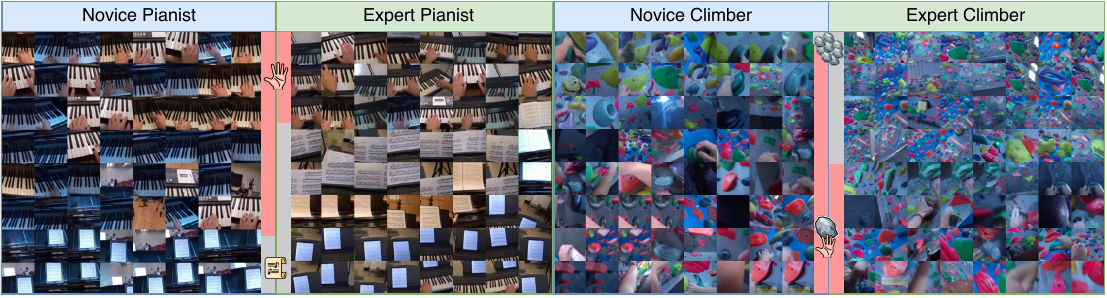}
    % \fbox{\rule{0pt}{2in} \rule{\textwidth}{0pt}}
    \vspace{-0.25in}
    \caption{\textbf{What does an expert vs.~novice tend to see more of?}
        In these distributions, each patch crops the egocentric frame based on the subject's gaze coordinates. Our representation surfaces interesting patterns, like (left two boxes) how   
    novice pianists fixate on their hands more often than experts do (77\% vs.~45\%, as quantified with hand detection), or (right two boxes) how bouldering experts exhibit greater gaze depth (1.4 m vs.~1.1 m) as they analyze moves further up the wall, resulting in smaller rocks in the crops.   These patterns emerging from in-the-wild video are consistent with and even deepen prior findings from psychology~\cite{psy_eyespan}. 
    }
    \label{fig:crop_analysis}
    \vspace{-0.4cm}
\end{figure*}

\subsection{Teacher model: Skill from action and attention} %Combining visual information with gaze}
\label{sec:teacher}

%\KGnote{should we first state the basics, that we will be training a %neural classifier....} 
First we train a classifier that takes both egocentric video (\emph{what the subject is doing}) and gaze  (\emph{how they are attending to their surroundings}) for skill level classification.
To ensure robust generalization across dynamic and static scenarios, SkillSight-T integrates gaze and visual signals through three components: (1) the interaction between the subject’s actions and gaze regions by applying the gaze attention to the visual encoder; (2) the sequence of subject's attended objects by encoding the gaze-cropped images; and (3) the dynamics of the subject’s gaze over time. Fig.~\ref{fig:model} shows the overview, and each part is described next.

\vspace*{-0.15in}
\paragraph{Action and gaze interaction} We leverage $g_{2d}^t$ to identify the gaze-attended region in $v^t$, and incorporate gaze information into the visual encoder $f_V$ (e.g. TimeSformer~\cite{Timesformer}). By knowing where the subject is looking, the model learns skill assessment by capturing the correlations between visual focus and actions. Specifically, we introduce a gaze-induced attention map $A_g=\{A_g^t\}_{t=1}^T$ into the first spatial encoder $f_{V,0}$ of $f_V$. Let $X=\{X^t\}_{t=1}^T$ be the input of $f_{V,0}$. For each timestep $t$, $f_{V,0}$ spatially divides $X^t$ into $p^2$ patches with size $L\times L$ and computes an attention map $A_v^t\in R^{p\times p}$. Next, we apply a Gaussian kernel centered at patch $c^t=\lfloor g_{2d}^t/L \rfloor$ and construct $A_g^t$ with:
\begin{equation}
    A_g^t[m,n] = 
\exp\!\left(-\tfrac{d_c^t(m,n)}{2\sigma^{2}}\right)/
\sum\limits_{m',n'} \exp\!\left(-\tfrac{d_c^t(m',n')}{2\sigma^{2}}\right),
\end{equation}
with $d_c^t(m,n) = ||(m,n) - c^t||^2$. The modified attention map is 
\begin{equation}
    A_m^t=\sigma(A_v^t+\lambda_c A_g^t),
\end{equation}
where $\sigma$ is the softmax operation and $\lambda_c$ is a learnable parameter for each scenario $c$, e.g., basketball, soccer. Finally, we obtain the embedding 
\begin{equation}
e_v=f_V(V,g_{2d}).
\end{equation}
Unlike prior gaze-based action recognition methods~\cite{EyeBeholder, MCN, gaze_att_i3d}, which pool gaze information at late-stage features, our method emphasizes gaze in the earliest spatial encoder, allowing the model to semantically highlight gaze regions. %\KGnote{did we ever ablate this / compare?} \JWnote{We haven't, but others methods are reasonable only in small segments. Will add experiments to supp.}

\vspace*{-0.15in}
\paragraph{Attended object sequence}
We represent the subject's attended objects by spatially cropping $v^t$ with $g^t_{2d}$. We observe that the distribution of attended objects for novices and experts differs significantly between tasks (see Fig.~\ref{fig:crop_analysis}). For instance, novice pianists fixate on their hands more often than expert pianists, who dwell more on the sheet music. This observation motivates leveraging the sequence of gazed-upon objects to reflect skill. 

While the sequence of attended objects is meaningful for skill assessment, we do not treat gaze-cropped image sequences $V_c=\{v_c^t\}_{t=1}^T$ as video \cite{egoexolearn} since the crops are taken from varying regions and lack spatial alignment across frames. Instead, we first compute semantic embeddings for $v_c^t$ using a pretrained image encoder $f_I$, and a subsequent temporal encoder $f_{T}$ models the sequence-level relationships, yielding the gaze-crop encoding:%. Mathematically, 
\begin{equation}
    e_c=f_T([f_I(v_c^1), ...f_I(v_c^T)]).
\end{equation}
\paragraph{Gaze dynamics}
While 2D gaze and ego-view video \cite{egoexolearn, EyeBeholder,MCN,gaze_att_i3d} highlight what a subject is looking at, they do not explicitly reflect the gaze dynamics such as the fixation frequency, the saccade speed, and the change of gaze location in the 3D environment---which all show significant differences across subjects with different skill levels \cite{psy_esports, psy_gazesport}. To that end, $G_i$ contains rich 3D information about the trajectory of the subject, the gaze direction, and the gaze depth. We encode $G_i$ using a transformer-based encoder $f_g$. To avoid bias in the gaze signals such as where the subject is facing, we normalize by calculating the gaze signals relative to the signals in the first frame. See Supp.~for details and analysis. Formally, this yields our third component to encode the gaze dynamics:
\begin{equation}
e_g = f_g(G).
\end{equation} 
We concatenate the features from the three components and pass the combined feature to the fusion layer $f_m$ for prediction. Specifically, we construct SkillSight-T as
\begin{equation}
    \hat{S}=\mathcal{F}_{v}(V,G) = f_m([e_{v}, e_{c}, e_{g}]),
\end{equation}
and use standard cross-entropy loss $L_{CE}$ for training. \JW{Our modules reason about where and why the user is looking by explicitly modeling the spatial and semantic interaction between gaze and visual, capturing skill-related patterns more effectively than simply inputting raw gaze \KG{(see Supp.)}.} 
% \KGnote{should this be ``more effectively than simply inputting raw gaze"? confusing because we are also concatenating (our) features.}

%\KGnote{around here, reader may ask ``why do you have to do all this massaging of the gaze inputs, use of crops, etc. vs. simpler end to end?  CHECK : Do we ever try a naive end-to-end baseline in results?  we should / and foreshadow here.}

\subsection{Student model: Distillation with gaze}\label{sec:distill}

Having defined the variant of our model that processes both gaze and video, next we generalize our approach to accommodate gaze alone---reducing power use and increasing privacy---without losing action-specific cues in video.

%To improve power efficiency, prior work has explored lowering the camera's frame rate \cite{egodistill} or selectively activating it only when needed \cite{chat2map, listentolook, egotrigger}. Nevertheless, these approaches require frequent camera activation which introduces nontrivial challenges as discussed in Section~\ref{sec:intro}.

To this end, we propose SkillSight-S, a lightweight method that relies solely on gaze for skill assessment. With only gaze signals required at inference, the egocentric camera remains deactivated. As already discussed, cognitive science establishes a strong correlation between gaze behavior and skill level~\cite{music_eyehand, psy_esports, psy_gazesport, psy_GazeHotspot}.  Furthermore, eye-tracking cameras consume far less power \cite{electrasight,power_numbers} than typical RGB cameras and mitigate privacy concerns since they only capture the user’s eyes rather than the full environment.  These properties make gaze a natural choice for power-efficient skill assessment.

%%% here I'm trying to get at why distillation should be possible.
But to what extent can video cues (what the user sees) be embedded \emph{into} the gaze signal?  Intuitively, people exhibit 
%Moreover, humans tend to exhibit 
consistent gaze patterns when observing certain objects or performing specific actions, making it natural to distill visual information into gaze. This correlation is amplified in the skill assessment setting, where the subject's actions are aligned with the goal of the skilled activity (e.g., cooking a dish, shooting a free throw), take place in skill-conducive environments (e.g., a kitchen, gym), and involve interactions with specific skill-relevant objects (e.g., pot and whisk, basketball and hoop).  These properties make our problem amenable to knowledge distillation.

%\KGnote{let's also be clear our idea is to imbue the gaze with the video/action/environment content, via distillation, as opposed to replacing video with gaze alone.  in general we could make a bigger deal about this in intro and again here --- the concept of distilling from video into gaze, which may be unexpected/interesting.}
%\JWnote{should we also mention it's good for privacy concern?} %\KGnote{yes.}

SkillSight-S consists of a transformer-based encoder, $f_s$ that takes $G$ as input. We train $f_s$ using knowledge distillation from the teacher $\mathcal{F}_v$ described in Sec.~\ref{sec:teacher}. We employ a distillation token, $t_{dis}$ \cite{deit}, to align the student features with those of the teacher. We also introduce an action recognition token, $t_{act}$, to classify the subject’s subtask, e.g. dribbling, and penalty kick, based on $G$.  This multi-branch architecture improves skill assessment by associating skilled gaze patterns with the subject's action. Specifically,
\begin{equation}
    \hat{e}_{s}, \hat{S}, \hat{a} = f_s([t_{cls}, t_{dis}, t_{act}, G])
\end{equation} 
where $\hat{a}$ predicts the subtask label, $\hat{S}$ predicts the skill level, and $\hat{e}_{s}$ is for distillation learning. The training objective of action classification is standard cross-entropy loss $L_{act}$. The distillation loss is computed as:
\begin{equation}
    L_{dis} = ||f_p(\hat{e}_{s})-f_t([e_{v}, e_{c}, e_{g}])||_1
\end{equation}
where $f_p$ is a projection layer that aligns the features of $\mathcal{F}_g$ and $\mathcal{F}_v$, a common practice in knowledge distillation \cite{fitnets, distilling}. We add another projection layer $f_t$ to mitigate the impact of modality-specific teacher signals that the student cannot effectively capture. We set the loss weights $\lambda_{dis}$ and $\lambda_{act}$ with validation data and train the student model with:
\begin{equation}
    L_{student} = L_{CE}+\lambda_{dis}L_{dis} + \lambda_{act}L_{act}.
\end{equation}
% with $\lambda_{dis}$ and $\lambda_{act}$ as loss weights set with validation data.

\begin{table*}[t]\footnotesize
    \centering
    \begin{tabular}{L{2.2cm}C{1.2cm}C{1.2cm}C{1.0cm}C{1.0cm}C{1.0cm}C{1.0cm}C{1.0cm}C{1.0cm}C{1.0cm}|C{1.0cm}}
        \toprule
        \multirow{2}{*}{Method} & 
        \multirow{2}{*}{Modalities} & 
        \multirow{2}{*}{Power (mW)} & 
        \multicolumn{7}{c|}{\textbf{EgoExo4D}~\cite{egoexo4d}} & 
        \multicolumn{1}{c}{\textbf{MSB}~\cite{badminton}} \\
        \cmidrule(lr){4-11} 
        & & & Overall & Soccer & Basketball & Bouldering & Music & Dance & Cooking & Badminton \\
        \midrule
        Majority vote & — & — & 32.3 & 74.4 & 35.7 & 0.0 & 44.0 & 43.3 & 50.9 & 41.1 \\
        E2GoMotion \cite{e2gomotion} & V & 329.3 & 34.9 & 55.8 & 49.0 & 3.0 & 16.7 & 50.4 & 50.9 & 43.5 \\
        Skillformer \cite{skillformer} & V & 697.5 & 42.4 & 74.4 & 42.0 & 27.0 & 47.2 & 43.3 & \textbf{58.5} & 44.0 \\
        TimeSformer \cite{Timesformer} & V & 697.5 & 45.5 & 76.7 & 53.2 & 28.0 & 36.1 & 44.8 & 56.6 & 50.5 \\
        EgoExoLearn \cite{egoexolearn} & V+G & 141.4 & 42.3 & 74.4 & 46.9 & 25.2 & 44.4 & 43.3 & 50.9 & 31.7 \\
        Beholder \cite{EyeBeholder} & V+G & 132.4 & 34.1 & 72.1 & 42.7 & 21.4 & 50.0 & 26.8 & 24.5 & 30.6 \\
        \textbf{SkillSight-T} & V+G & 943 & \textbf{50.1} & \textbf{81.4} & \textbf{55.2} & \textbf{28.9} & \textbf{50.0} & \textbf{56.7} & \textbf{58.5} & \textbf{53.1} \\
        \midrule
        X3D-XS \cite{X3d} & V & 88 & 34.2 & 72.1 & \textbf{45.5} & 24.5 & 38.9 & 26.8 & 17.0 & 42.7 \\
        EgoDistill \cite{egodistill} & V+I & 16.5 & \underline{42.6} & 74.4 & 35.0 & \textbf{38.4} & \underline{50.0} & \underline{43.3} & \underline{43.4} & \underline{43.4} \\
        EgoTrigger \cite{egotrigger} & V+A & 9.9 & 34.1 & 65.1 & 37.8 & 22.6 & 41.7 & 26.8 & 5.7 & \textit{no audio}  \\
        Gaze-only & G & 9.5 & 37.0 & \underline{76.7} & 25.2 & 31.5 & 44.4 & 40.2 & 39.6 & 42.3 \\
        \textbf{SkillSight-S} & G & 9.5 & \textbf{44.4} & \textbf{79.1} & \underline{42.0} & \underline{34.6} & \textbf{52.8} & \textbf{44.1} & \textbf{47.2} & \textbf{47.0} \\
        \bottomrule
    \end{tabular}
    \caption{\textbf{Results on the Ego-Exo4D \cite{egoexo4d} (left) and Multi-Sense Badminton (MSB) \cite{badminton} (right) benchmarks.} \textbf{Top section:} SkillSight-T outperforms all prior methods across all scenarios in terms of accuracy (\%). \textbf{Bottom section:} SkillSight-S surpasses all power-efficient methods in overall accuracy (44.4\%) as well as 5 of the 7 individual scenarios. Even when compared to the more expensive, power-consuming baselines (top section), SkillSight-S still ranks second in overall accuracy, while using \JW{14}$\times$ to 73$\times$ less power (mW).  Bold face indicates best accuracy and underline indicates second-best. \JW{(V:Visual, G:Gaze, I:IMU, A:Audio).}}\label{tab:egoexo4d_result}
    \vspace{-0.4cm}
\end{table*}

\begin{table}[t]\footnotesize
    \centering
    \begin{tabular}{L{2.5cm}C{1.5cm}C{1.5cm}}
        \toprule
        \multicolumn{3}{c}{\textbf{Expert–Novice Soccer}~\cite{expert_novice_soccer}} \\
        \midrule
        Method & Modalities & Overall \\
        \midrule
        Majority vote & — & 50.0 \\
        Gaze-only & G & 66.0 \\
        Body-motion-only & M & 71.2 \\
        \midrule
        \textbf{SkillSight-S} & G & \underline{72.6} \\
        Body-motion+Gaze & M+G & \textbf{73.3} \\
        \bottomrule
    \end{tabular}
    \vspace{-0.2cm}
    \caption{\textbf{Results on Expert-Novice Soccer~\cite{expert_novice_soccer}.} Since the Expert-Novice Soccer does not include video, we use transformer baselines with full \KG{body} \JW{motion (M) and eye-tracking gaze (G).} SkillSight-S outperforms both the gaze-only and motion-only baselines, showing the effectiveness of our distillation technique.}
    \label{tab:soccer_result}
    \vspace{-0.6cm}
\end{table}

\subsection{Data and implementation details}
\label{sec:data_detail}

\paragraph{Method and training details} Following the Ego-Exo4D benchmark \cite{egoexo4d}, we segment long videos into 10 equally spaced clips and average segment-level predictions for classification. %The benchmark’s default 1-second clip duration is insufficient to capture complete action sequences. 
\JW{Note that we use untrimmed videos without making strong assumptions about where the skilled portions of the sequence occur.}
%\KGnote{should we point out that we are not making strong assumptions about where the skilled component of the sequence is, i.e., our inputs are untrimmed?}
To better model skill-relevant dynamics, we configure both the teacher and student models to process 16-frame clips at 2 FPS, balancing temporal coverage with computational efficiency.  We use TimeSformer~\cite{Timesformer} pretrained on EgoVLPv2~\cite{egovlpv2} as $f_V$, achieving state-of-the-art egocentric video understanding, and Dinov2~\cite{dinov2} as $f_I$ for its strong spatial representation. Both $f_s$ and $f_g$ are 4-layer transformer encoders with a 768-dimensional hidden size, and $f_m$ is a 3-layer MLP. SkillSight-T is trained for 15 epochs using SGD (learning rate $5\times10^{-3}$, batch size 8), and SkillSight-S for 10 epochs using AdamW (learning rate $1\times10^{-4}$, batch size 32). All models are trained on 8 NVIDIA Quadro RTX 6000 GPUs. SkillSight-S processes a single sample in 1.6 ms on average using a single GPU. 
%\KGnote{here or later, should we report actual run time at inference?} 

\vspace*{-0.2in}
\paragraph{Data sources and statistics} 
We evaluate our method on three datasets.  (1) \textbf{Ego-Exo4D} \cite{egoexo4d} consists of 5,048 videos recorded by 740 participants. 
We use all the scenarios provided with the demonstrator proficiency estimation benchmark: soccer, basketball, rock climbing,  dance, %and static scenarios, e.g., 
music, and cooking. Following \cite{egoppg}, we use 10\% from the official training set for validation, and the held-out official validation set for testing. Each subject is annotated with one of four skill levels: novice, early expert, intermediate expert, and late expert. (2) \textbf{Multi-Sense Badminton} \KG{(MSB)} \cite{badminton} encompasses 7,763 badminton forehand and backhand swing data from 25 players. The skill levels are annotated into beginner, intermediate, and expert. We follow the official cross-validation split. (3) \textbf{Expert-Novice Soccer} \cite{expert_novice_soccer} contains 288 recordings from 8 subjects performing 9 different soccer movements such as kicks, dribbling, and juggling. Subjects are labeled as expert and novice. We follow the official cross-validation.

These datasets were chosen because they have gaze, camera pose, and ground truth skill labels provided by expert annotators (e.g., domain-specific coaches and teachers). In total, the gaze is from 3 distinct wearable devices, reflecting today's good availability of this modality.
%All datasets provide gaze signals, as well as glass rotation and translation. 
Ego-Exo4D and Expert-Novice Soccer include 3D gaze, while %Multi-Sense Badminton 
MSB provides 2D gaze. Expert-Novice Soccer does not contain video; therefore, we train its teacher model using \KG{body} motion (21 joint positions over time) and gaze. %Proficiency ratings for all datasets are provided by expert annotators. 
For all datasets, no subject overlaps between the train-test splits.

\section{Experiment}\label{sec:expts}

\begin{figure*}[t]
    \centering
    % Placeholder box (adjust height as needed)
    % \fbox{\rule{0pt}{5in} \rule{\textwidth}{0pt}}
    \includegraphics[width=\textwidth]{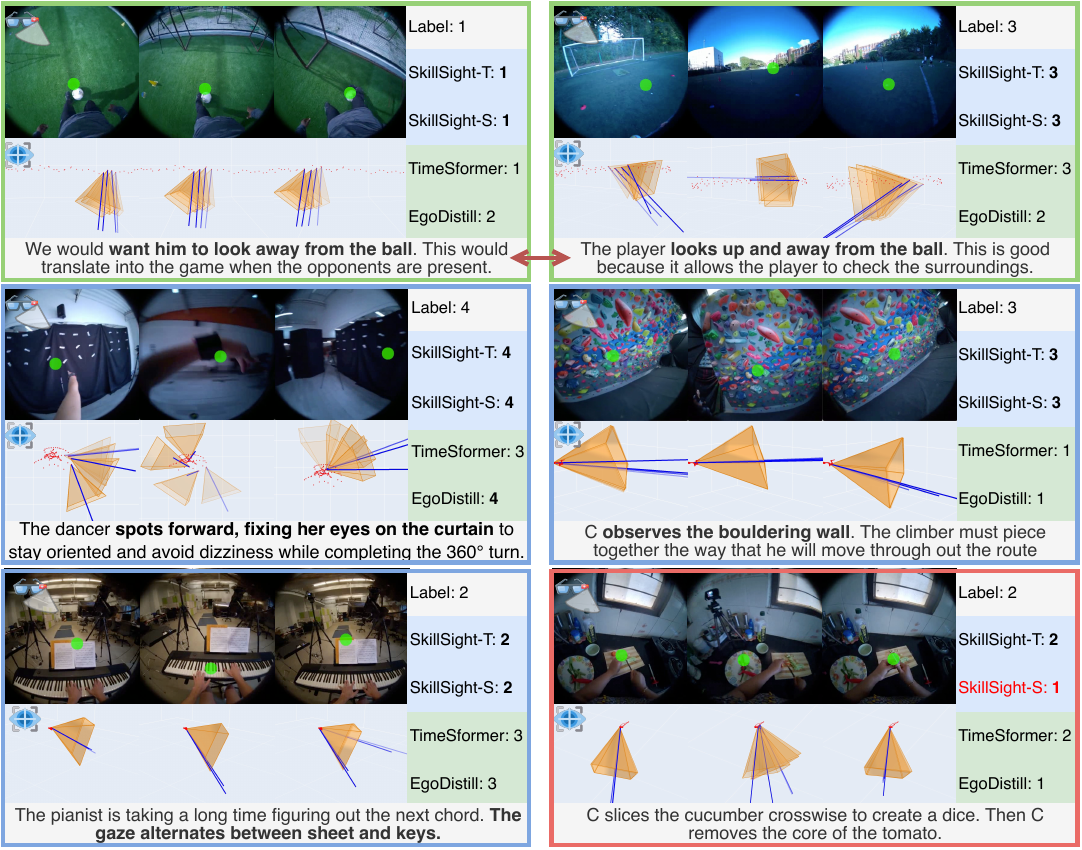}
    \vspace{-0.25in}
    \caption{\textbf{Qualitative results.} Both SkillSight-T and SkillSight-S better predict skill level than prior work. Experts and novices show distinct gaze patterns consistent with Ego-Exo4D~\cite{egoexo4d} expert commentaries, shown for reference but not used by any model. The last example (bottom right) shows a failure case, highlighting the challenge of assessing skill from subtle movements. %\KGnote{here and in Fig 1, how to understand the blue gaze bars and shading.} 
    Blue rays show gaze direction and depth, and \KG{frustrum/ray} shading indicates recent glasses motion. Ground-truth labels range from 1 (novice) to 4 (late expert).}
    \label{fig:qualitative}
    \vspace{-0.3cm}
\end{figure*}

\begin{figure}[t] % [t] = top of column, can also use [h] or [b]
    \centering
    \includegraphics[width=\linewidth]{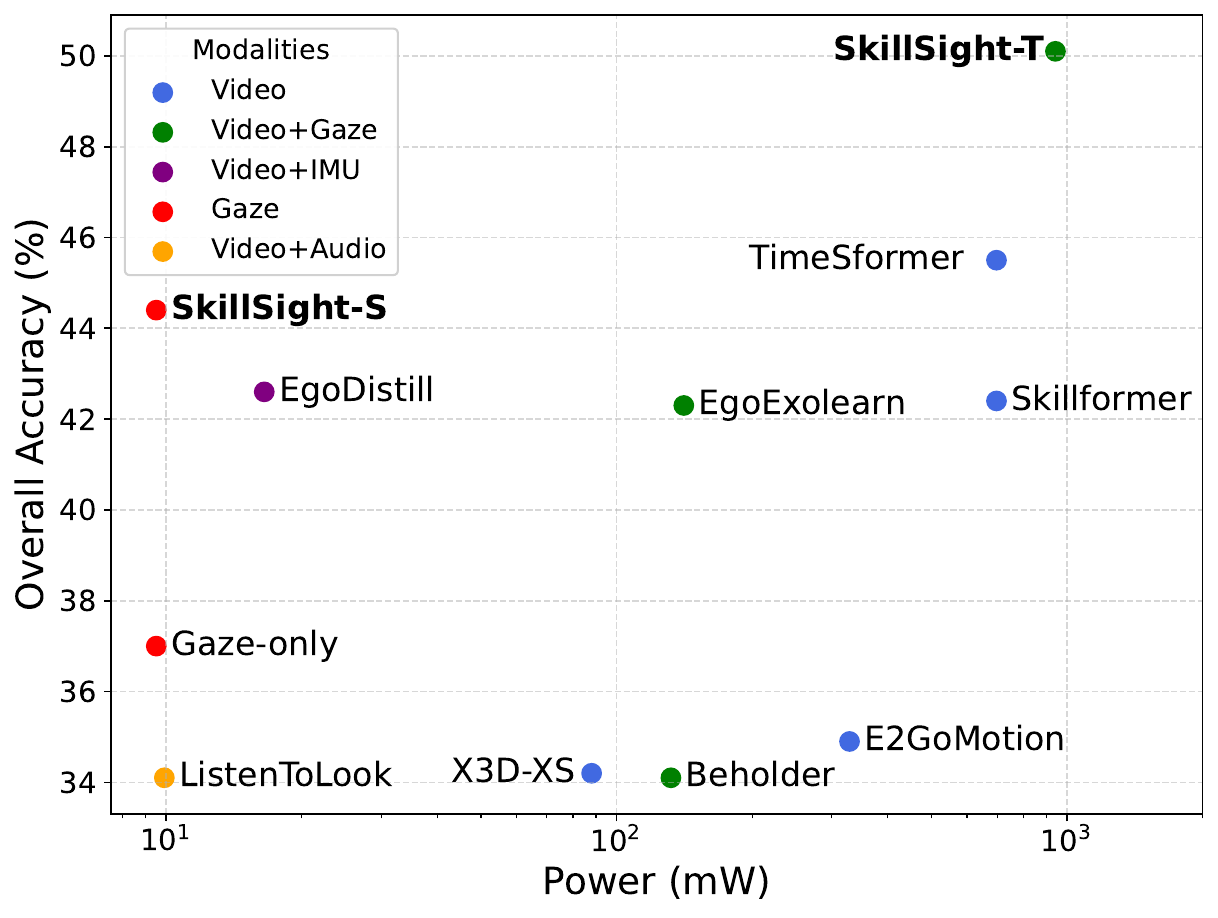}
    % Placeholder box
    % \fbox{\rule{0pt}{1.5in} \rule{0.9\columnwidth}{0pt}}
    % Caption
    \vspace{-0.8cm}
    \caption{\textbf{Power–accuracy tradeoff}. SkillSight-T outperforms all baselines in accuracy, while SkillSight-S achieves the second-best accuracy and consumes the least energy. The optimal method would attain maximal accuracy with minimal power (top left). 
    % \KGnote{i don't think we define H, A, G anywhere.}
    }
    \label{fig:power}
    \vspace{-0.4cm}
\end{figure}

We first describe baselines, followed by results and qualitative examples. Finally, we analyze power efficiency and our performance across different scenarios.

\vspace*{-0.15in}
\paragraph{Baselines} We compare to video action\JW{/skill} recognition methods~\cite{Timesformer, X3d, skillformer}, methods using diverse modalities from glasses~\cite{egodistill,egotrigger,e2gomotion}, and ego methods using gaze~\cite{egoexolearn,EyeBeholder}:
\begin{itemize}
\item \textbf{TimeSformer} \cite{Timesformer}, \textbf{X3D-XS} \cite{X3d}, \textbf{Skillformer} \cite{skillformer}: The first two are standard video-classification models. X3D-XS is an efficient architecture suitable for deployment on smart glasses, while TimeSformer represents the Ego-Exo4D baseline for proficiency estimation \cite{egoexo4d}. Skillformer builds on TimeSformer, fine-tuned via LoRA \cite{lora}.
\item \textbf{EgoDistill} \cite{egodistill}, \textbf{EgoTrigger} \cite{egotrigger}: EgoDistill is a power-efficient approach that processes a single RGB frame together with the corresponding sequence of IMU readings from the glasses for action recognition. EgoTrigger, similar to \cite{listentolook}, reduces power consumption by leveraging audio cues to decide whether to process the visual stream.
\item \textbf{E2GoMotion} \cite{e2gomotion}: The method leverages event-camera data for action recognition. Since no skill dataset contains event-camera recordings, we provide full-frame-rate optical flow to their model as a proxy, identical to % This represents an 
the upper bound %for their method as 
reported in their study.
\item \textbf{EgoExoLearn}~\cite{egoexolearn}, \textbf{Beholder}~\cite{EyeBeholder}: EgoExoLearn crops ego-view video around gaze points and uses I3D~\cite{i3d} for skill classification, while Beholder performs gaze-weighted pooling of visual features for action recognition.
\item \textbf{Gaze-only}: This method only takes gaze as input and shares the same architecture with SkillSight-S. We use cross-entropy loss for training without distillation.
\end{itemize}

%\KGnote{let's have a summary statement here pointing out that only one of these baselines studied gaze for skill assessment(egoexolearn) and the others were considered for action recognition, yet we're seeing if a straighforward adaptation of them can solve this task.  Also in results be sure to help reviewer keep straight when we're using video or not and same for baselines.}

Of all the baselines, only Skillformer~\cite{skillformer} and EgoExoLearn~\cite{egoexolearn} are \KG{specifically} for skill assessment, and only EgoExoLearn utilizes gaze. Other models~\cite{Timesformer, X3d, egodistill, egotrigger, e2gomotion, EyeBeholder} originally target action recognition; to broaden the pool of baselines, we adapt them for skill assessment by adjusting the output dimension and training on the same datasets.
X3D-XS~\cite{X3d}, EgoDistill~\cite{egodistill}, and EgoTrigger~\cite{egotrigger} are power-efficient methods leveraging less computation or lightweight modalities. 
We evaluate using standard accuracy metrics and estimated power consumption.
%KGnote{insert the clusters of refs here too.}

% \textbf{Ablations.} In addition to the strong baselines, we evaluate ablated variants to isolate each component’s contribution. \textbf{SkillSight-T w/o gaze-att}, \textbf{SkillSight-T w/o crop}, and \textbf{SkillSight-T w/o gaze-seq} remove, $f_V$, $(f_{Temp},f_I)$, and $f_g$, respectively from \textbf{SkillSight-T}). \textbf{Gaze-Only} is trained using only gaze as input without distillation.

\vspace*{-0.15in}
\paragraph{Results}
 Table \ref{tab:egoexo4d_result} reports results on Ego-Exo4D~\cite{egoexo4d} and Multisense Badminton~\cite{badminton}. SkillSight-T outperforms all baselines across seven scenarios in both datasets, achieving an average relative gain of 10\% over the strongest baseline. %\KGnote{consider making some punchline about relative gains too?} 
 Remarkably, SkillSight-S, which uses only gaze as input, outperforms not only all the power-efficient baselines (bottom), but also the majority of the power-hungry baselines---despite using 14$\times$ to 73$\times$ less power (details below).  It also achieves the best performance among power-efficient baselines in five of seven individual scenarios.\footnote{\JWCam{EgoPPG~\cite{egoppg} reports its performance on a modified EgoExo4D test set, which is not directly comparable to the results in Tab.~\ref{tab:egoexo4d_result} . On the modified test set, SkillSight-T outperforms EgoPPG by 11\% relative. SkillSight-S exceeds EgoPPG by 0.5\% relative and uses significantly less power.}}

Notably, SkillSight-T is superior in both static scenes, i.e. cooking and music, and dynamic sports, i.e. soccer, basketball, dance, rock climbing, dancing, and badminton. We attribute the robust prediction to our designs for incorporating gaze with vision, allowing the model to learn from the attended objects, the actions, and the gaze transition. \JW{We show that SkillSight-T outperforms a naive end-to-end model by %a relative gain of 
$8\%$ as well as more \textbf{ablations} in Supp.}

Despite having the lowest power consumption, SkillSight-S outperforms other power-efficient baselines \KG{(Tab.~\ref{tab:egoexo4d_result}, bottom)}. We see a significant improvement compared to the Gaze-only baseline. This shows that SkillSight-S effectively learns the knowledge of SkillSight-T through our distillation technique. Models that rely only on first-person visual input, e.g., X3D-XS~\cite{X3d}, fail to learn consistent skill patterns across scenarios. EgoDistill~\cite{egodistill} and EgoTrigger~\cite{egotrigger} use a single frame together with head rotation or audio to represent the subject’s action; however, these modalities struggle to reveal subtle differences in actions for rating skill. On the other hand, gaze directly captures how subjects actively shift attention to complete tasks. This highlights gaze as a compact, highly informative signal for low-power skill assessment. 

We present qualitative results in Fig.~\ref{fig:qualitative}. We see that across different scenarios, experts and novices demonstrate different gaze patterns. For instance, when dribbling in soccer (first row), the novice looks down on the ball while the expert looks away from the ball to check the surroundings. When expert dancers perform a spin (middle left), they fixate their eyes early to the front to avoid dizziness. These patterns show important cues that our methods leverage to access skills robustly, showing the benefit of explicitly modeling multiple aspects of gaze and skill together. By contrast, TimeSformer~\cite{Timesformer} and Skillformer~\cite{skillformer}—neither of which uses gaze—struggle when subjects exhibit few motion cues. For example, an ego-view clip alone may not reveal that a performer shifts gaze from sheet music to their hands (bottom left), offering limited cues for skill assessment. EgoExoLearn~\cite{egoexolearn} and Beholder~\cite{EyeBeholder} restrict processing to visual regions around gaze. While this approach is effective when gaze remains on the hands, it discards valuable contextual information when gaze shifts away from the body. For example, in bouldering (middle right), they may focus on the wall. Prior approaches that limit attention to the gaze region therefore overlook cues critical for assessing skill. Finally, we show a failure case where gaze does not reflect skill when the subject is slicing vegetables (bottom right), showcasing the limitation of %only using 
gaze when subtle hand movements are required.
% The result on MultiSense Badminton~\cite{badminton} (Table ~\ref{tab:egoexo4d_result} right) further supports this: both methods perform poorly on the badminton scenario, where the player’s attention is on the ball but correct assessment depends on the relationship between player motion and ball trajectory. See Supp. 

Table~\ref{tab:soccer_result} 
reports results on Expert-Novice Soccer~\cite{expert_novice_soccer}.  They
highlight the effectiveness of our distillation framework. SkillSight-S, using only smart-glasses signals, surpasses both Gaze-only and Body-motion-only baselines, \JW{the latter of which requires subjects to wear body-mounted IMUs.} %\KGnote{not clear to me why this IMU is costly setup?}  
Across all datasets, our method enhances gaze-based models and achieves competitive, power-efficient performance suitable for skill assessment on smart glasses.

\vspace*{-0.15in}
\paragraph{Efficiency analysis.} 
Accurately measuring power consumption for wearable device applications is crucial. 
\JWnew{Using well-established measurements \cite{egoexo4d}}, the total energy consumption can be divided into three components: 
sensor triggering energy~($\gamma$), compute energy~($\alpha$), and memory transfer energy~($\beta$). 
See Supp.~for full explanation. 
We employ weighting parameters based on real-world estimates of the power consumption. 
Specifically, $\alpha = 4.6~\mathrm{pJ}/\mathrm{MAC}$ \cite{power_alpha}, 
$\beta = 80~\mathrm{pJ}/\mathrm{byte}$ \cite{power_beta}, 
$\gamma_{\mathrm{rgb}} = 35~\mathrm{mW}$, 
$\gamma_{\mathrm{IMU}} = 1.2~\mathrm{mW}$, 
$\gamma_{\mathrm{audio}} = 0.3~\mathrm{mW}$ \cite{power_numbers}, 
and $\gamma_{\mathrm{eye}} = 7.8~\mathrm{mW}$ \cite{electrasight}. 
All values are taken from hardware designed for smart glasses. 

The overall energy consumption rate of a model is:
\begin{equation}
    P = \frac{\alpha N}{T} + \frac{\beta B}{T} + \sum_m \gamma_m\delta_m,
\end{equation}
where $N$ is the number of MACs in the model forward pass, $B$ is the number of bytes required for read/write operations, $m$ indexes the modalities used by the model, $\delta_m=1$ when the model uses modality $m$, and $0$ otherwise. $T$ is the time interval between successive inferences.  

Figure~\ref{fig:power} shows that SkillSight-S achieves the best overall trade-off between power consumption and accuracy. It outperforms all power-efficient baselines while reducing the power consumption of the best baseline, i.e. EgoDistill~\cite{egodistill}, by 43$\%$. Moreover, SkillSight-S demonstrates competitive performance compared to video-based methods, which are power intensive \emph{regardless of the architecture} due to the energy cost of sensing and visual feature encoding. Compared to TimeSformer~\cite{Timesformer}, SkillSight-S achieves over 
$73\times$ lower energy cost
% \KGnote{check.} 
with only a $1.1\%$ drop in accuracy. Our approach provides an efficient foundation for real-time assistance or skill assessment.
%\KGnote{see group meeting notes about explaining backbone contribution to power.}
\begin{figure}[t] % [t] = top of column, can also use [h] or [b]
    \centering
    \includegraphics[width=\linewidth]{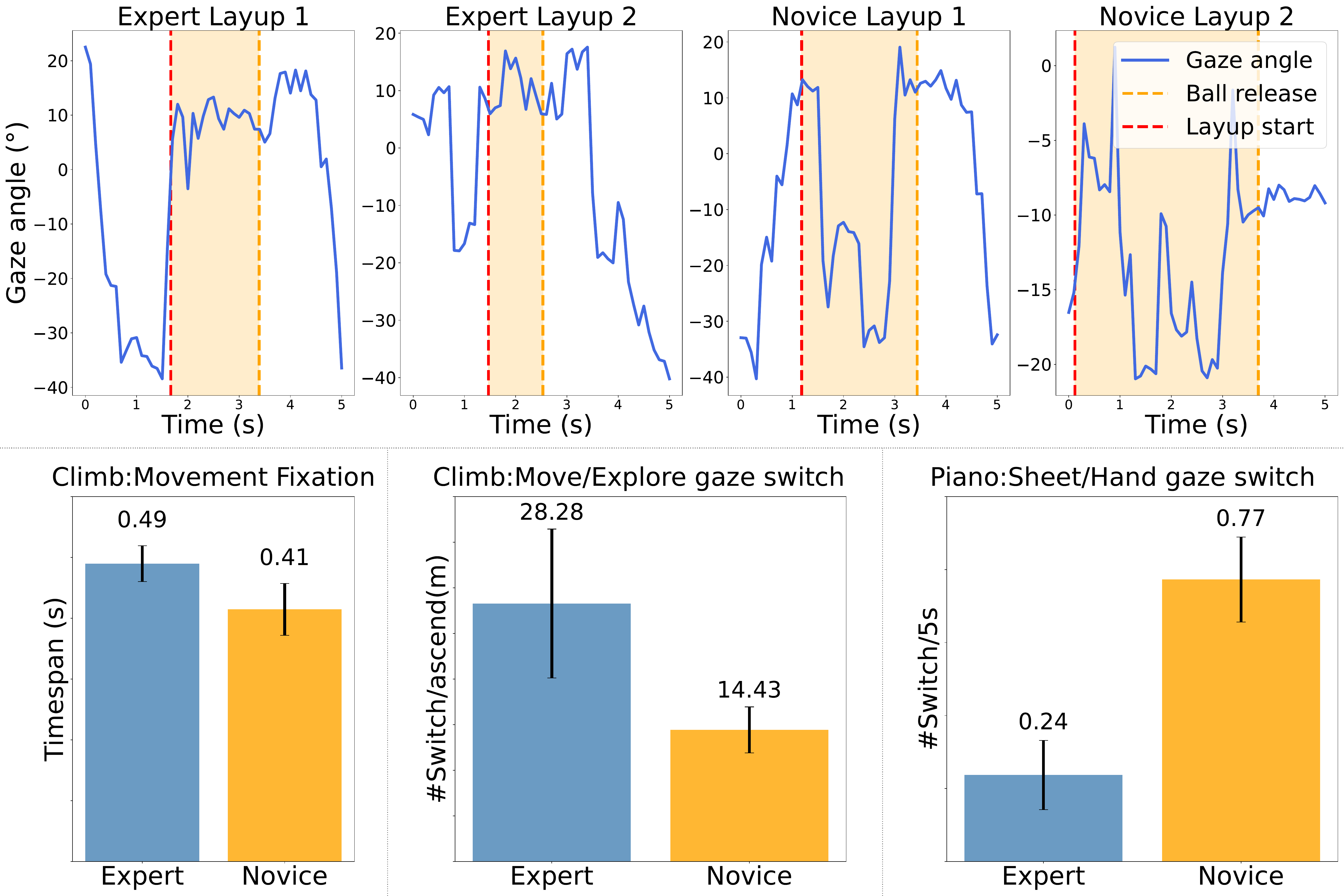}
    % Placeholder box
    % \fbox{\rule{0pt}{1.5in} \rule{0.9\columnwidth}{0pt}}
    % Caption
    \vspace{-0.7cm}
    \caption{\textbf{Gaze pattern analysis.} SkillSight-S reveals distinct gaze patterns between model-predicted experts and novices.}
    \label{fig:analysis}
    \vspace{-0.3cm}
\end{figure}
\vspace*{-0.15in}
\paragraph{Psychology insight from SkillSight.}\JW{
Figures~\ref{fig:analysis} \KG{and~\ref{fig:crop_analysis}} show gaze behavior insights from SkillSight-S. In basketball layups, model-predicted experts consistently look up toward the rim, while novices look down at the ball (top). In bouldering, our predicted experts show longer movement-related fixations (e.g., grasp or foot placement) (bottom left), consistent with sports science~\cite{rock_science}. Beyond that, experts switch more often between movement-related and exploratory fixations when ascending (bottom middle). Figure~\ref{fig:crop_analysis} shows that novice pianists focus more on the hands, aligning with psychology findings~\cite{psy_eyespan}, while SkillSight further shows more frequent gaze transitions between the sheet and hands (\KG{Fig.~\ref{fig:analysis}}, bottom right). SkillSight \KG{not only} aligns with established psychological findings, it also facilitates finer exploration of expert-novice gaze strategies.}

\vspace*{-0.05in}

\section{Conclusion}
\label{sec:conclusion}

%-------------------------------------------------------------------------
We investigate how gaze behavior reflects skill level across dynamic and static scenarios. Our methods integrate gaze with egocentric visuals to assess skill by modeling attention during task execution. Moreover, our distillation framework enables a lightweight model using only gaze, achieving competitive accuracy while using significantly less power. Our work lays the foundation for future AI-driven instructional and assistive systems on smart glasses.

\section*{Acknowledgement}
\label{sec:acknowledgement}
\JWCam{Research supported in part by a gift from Amazon and the UT Austin IFML NSF AI Institute. We thank Zihui Xue for valuable advice on head pose representation and normalization process, and the members of the UT Austin Computer Vision Group for helpful discussions.} 
{
    \small
    \bibliographystyle{ieeenat_fullname}
    \bibliography{main}

@String(CVPR= {IEEE Conf. Comput. Vis. Pattern Recog.})

@String(ECCV= {Eur. Conf. Comput. Vis.})

@String(BMVC= {Brit. Mach. Vis. Conf.})

@String(ICASSP=	{ICASSP})

@String(ICLR = {Int. Conf. Learn. Represent.})

@String(CVPR  = {CVPR})

@String(ECCV  = {ECCV})

@String(BMVC  =	{BMVC})

@String(ICLR  = {ICLR})

@article{egogaze,
  title={In the eye of mllm: Benchmarking egocentric video intent understanding with gaze-guided prompting},
  author={Peng, Taiying and Hua, Jiacheng and Liu, Miao and Lu, Feng},
  journal={arXiv preprint arXiv:2509.07447},
  year={2025}
}

@article{Vickers1996,
  author    = {Joan N. Vickers},
  title     = {Visual control when aiming at a far target},
  journal   = {Journal of Experimental Psychology: Human Perception and Performance},
  year      = {1996},
  volume    = {22},
  number    = {2},
  pages     = {342--354},
  doi       = {10.1037/0096-1523.22.2.342}
}

@article{Vickers2011Surgery,
  author    = {Vine, Samuel J. and Chaytor, R. J. and McGrath, J. S. and Masters, R. S. W.},
  title     = {Gaze training improves laparoscopic surgical performance},
  journal   = {Surgical Endoscopy},
  year      = {2011},
  volume    = {25},
  number    = {12},
  pages     = {3731--3739},
  doi       = {10.1007/s00464-011-1784-3}
}

@article{DraiZerbib2012,
  author    = {Veronique Drai-Zerbib and Emmanuel Baccino},
  title     = {The influence of expertise in music reading on the detection of temporal violations},
  journal   = {Visual Cognition},
  year      = {2012},
  volume    = {20},
  number    = {3},
  pages     = {267--282},
  doi       = {10.1080/13506285.2012.658366}
}

@article{Causer2014,
  author    = {Causer, Joe and Harvey, Adam and Snelgrove, Richard and Arsenault, Gary and Vartanian, Oshin},
  title     = {Quiet eye training improves surgical performance: A randomized controlled study},
  journal   = {Frontiers in Psychology},
  year      = {2014},
  volume    = {5},
  pages     = {821},
  doi       = {10.3389/fpsyg.2014.00821}
}

@article{Vickers2016Driving,
  author    = {Vickers, Joan N. and Lew, D. J.},
  title     = {Quiet eye duration predicts expertise in a simulated driving task},
  journal   = {Cognitive Processing},
  year      = {2016},
  volume    = {17},
  number    = {3},
  pages     = {311--319},
  doi       = {10.1007/s10339-016-0760-9}
}

@inproceedings{FitnessAQ,
  title={Domain knowledge-informed self-supervised representations for workout form assessment},
  author={Parmar, Paritosh and Gharat, Amol and Rhodin, Helge},
  booktitle={European Conference on Computer Vision},
  pages={105--123},
  year={2022},
  organization={Springer}
}

@inproceedings{Piano,
  title={Piano skills assessment},
  author={Parmar, Paritosh and Reddy, Jaiden and Morris, Brendan},
  booktitle={2021 IEEE 23rd international workshop on multimedia signal processing (MMSP)},
  pages={1--5},
  year={2021},
  organization={IEEE}
}

@inproceedings{Surgery,
  title={Towards accurate and interpretable surgical skill assessment: A video-based method incorporating recognized surgical gestures and skill levels},
  author={Wang, Tianyu and Wang, Yijie and Li, Mian},
  booktitle={International Conference on Medical Image Computing and Computer-Assisted Intervention},
  pages={668--678},
  year={2020},
  organization={Springer}
}

@article{skating,
  title={Learning to score figure skating sport videos},
  author={Xu, Chengming and Fu, Yanwei and Zhang, Bing and Chen, Zitian and Jiang, Yu-Gang and Xue, Xiangyang},
  journal={IEEE transactions on circuits and systems for video technology},
  volume={30},
  number={12},
  pages={4578--4590},
  year={2019},
  publisher={IEEE}
}

@inproceedings{diving,
  title={Finediving: A fine-grained dataset for procedure-aware action quality assessment},
  author={Xu, Jinglin and Rao, Yongming and Yu, Xumin and Chen, Guangyi and Zhou, Jie and Lu, Jiwen},
  booktitle={CVPR},
  pages={2949--2958},
  year={2022}
}

@inproceedings{egoexolearn,
  title={Egoexolearn: A dataset for bridging asynchronous ego-and exo-centric view of procedural activities in real world},
  author={Huang, Yifei and Chen, Guo and Xu, Jilan and Zhang, Mingfang and Yang, Lijin and Pei, Baoqi and Zhang, Hongjie and Dong, Lu and Wang, Yali and Wang, Limin and others},
  booktitle={Proceedings of the IEEE/CVF Conference on Computer Vision and Pattern Recognition},
  pages={22072--22086},
  year={2024}
}

@inproceedings{egoexo4d,
  title={Ego-exo4d: Understanding skilled human activity from first-and third-person perspectives},
  author={Grauman, Kristen and Westbury, Andrew and Torresani, Lorenzo and Kitani, Kris and Malik, Jitendra and Afouras, Triantafyllos and Ashutosh, Kumar and Baiyya, Vijay and Bansal, Siddhant and Boote, Bikram and others},
  booktitle={Proceedings of the IEEE/CVF Conference on Computer Vision and Pattern Recognition},
  pages={19383--19400},
  year={2024}
}

@inproceedings{AmIABaller,
  title={Am I a baller? Basketball performance assessment from first-person videos},
  author={Bertasius, Gedas and Soo Park, Hyun and Yu, Stella X and Shi, Jianbo},
  booktitle={Proceedings of the IEEE international conference on computer vision},
  pages={2177--2185},
  year={2017}
}

@inproceedings{narrativeAQA,
  title={Narrative action evaluation with prompt-guided multimodal interaction},
  author={Zhang, Shiyi and Bai, Sule and Chen, Guangyi and Chen, Lei and Lu, Jiwen and Wang, Junle and Tang, Yansong},
  booktitle={Proceedings of the IEEE/CVF Conference on Computer Vision and Pattern Recognition},
  pages={18430--18439},
  year={2024}
}

@inproceedings{videotextAQA2,
    title={RICA\textasciicircum 2: Rubric-Informed, Calibrated Assessment of Actions}, 
    author={Majeedi, Abrar and Gajjala, Viswanatha Reddy and Namburi, Satya Sai Srinath GNVV and Li, Yin},
    booktitle={Proceedings of the European Conference on Computer Vision (ECCV)},
    year={2024}
}

@inproceedings{videotextAQA3,
  title={Vision-language action knowledge learning for semantic-aware action quality assessment},
  author={Xu, Huangbiao and Ke, Xiao and Li, Yuezhou and Xu, Rui and Wu, Huanqi and Lin, Xiaofeng and Guo, Wenzhong},
  booktitle={European Conference on Computer Vision},
  pages={423--440},
  year={2024},
  organization={Springer}
}

@article{visualtextAQA,
  title={Visual-semantic alignment temporal parsing for action quality assessment},
  author={Gedamu, Kumie and Ji, Yanli and Yang, Yang and Shao, Jie and Shen, Heng Tao},
  journal={IEEE Transactions on Circuits and Systems for Video Technology},
  year={2024},
  publisher={IEEE}
}

@inproceedings{beatsAQA,
  title={From Beats to Scores: A Multi-Modal Framework for Comprehensive Figure Skating Assessment},
  author={Wang, Fengshun and Wang, Qiurui and Chen, Dan},
  booktitle={Proceedings of the Computer Vision and Pattern Recognition Conference},
  pages={5905--5914},
  year={2025}
}

@article{multimodalAQA,
  title={Multimodal action quality assessment},
  author={Zeng, Ling-An and Zheng, Wei-Shi},
  journal={IEEE Transactions on Image Processing},
  volume={33},
  pages={1600--1613},
  year={2024},
  publisher={IEEE}
}

@inproceedings{AV-AQA,
  title={Language-Guided Audio-Visual Learning for Long-Term Sports Assessment},
  author={Xu, Huangbiao and Ke, Xiao and Wu, Huanqi and Xu, Rui and Li, Yuezhou and Guo, Wenzhong},
  booktitle={Proceedings of the Computer Vision and Pattern Recognition Conference},
  pages={23967--23977},
  year={2025}
}

@inproceedings{fineparser,
  title={Fineparser: A fine-grained spatio-temporal action parser for human-centric action quality assessment},
  author={Xu, Jinglin and Yin, Sibo and Zhao, Guohao and Wang, Zishuo and Peng, Yuxin},
  booktitle={Proceedings of the IEEE/CVF Conference on computer vision and pattern recognition},
  pages={14628--14637},
  year={2024}
}

@inproceedings{rock_science,
  title={EXPLORING NEW HEIGHTS: VISUAL BEHAVIOUR OF NOVICE, INTERMEDIATE, AND EXPERIENCED CLIMBERS},
  author={Vansteenkiste, P and Zeuwts, L and Deconinck, FJA and Lenoir, M},
  booktitle={CONGRESS BOOK},
  pages={116},
  year={2018}
}

@article{skeleton,
  title={Lucidaction: A hierarchical and multi-model dataset for comprehensive action quality assessment},
  author={Dong, Linfeng and Wang, Wei and Qiao, Yu and Sun, Xiao},
  journal={Advances in Neural Information Processing Systems},
  volume={37},
  pages={96468--96482},
  year={2024}
}

@article{skeleton2,
  title={Multi-skeleton structures graph convolutional network for action quality assessment in long videos},
  author={Lei, Qing and Li, Huiying and Zhang, Hongbo and Du, Jixiang and Gao, Shangce},
  journal={Applied Intelligence},
  volume={53},
  number={19},
  pages={21692--21705},
  year={2023},
  publisher={Springer}
}

@article{skeleton3,
  title={Efficient and robust skeleton-based quality assessment and abnormality detection in human action performance},
  author={Elkholy, Amr and Hussein, Mohamed E and Gomaa, Walid and Damen, Dima and Saba, Emmanuel},
  journal={IEEE journal of biomedical and health informatics},
  volume={24},
  number={1},
  pages={280--291},
  year={2019},
  publisher={IEEE}
}

@inproceedings{gaze_action_recognition,
  title={Deep future gaze: Gaze anticipation on egocentric videos using adversarial networks},
  author={Zhang, Mengmi and Teck Ma, Keng and Hwee Lim, Joo and Zhao, Qi and Feng, Jiashi},
  booktitle={Proceedings of the IEEE conference on computer vision and pattern recognition},
  pages={4372--4381},
  year={2017}
}

@inproceedings{basketball_gaze,
  title={Predicting behaviors of basketball players from first person videos},
  author={Su, Shan and Pyo Hong, Jung and Shi, Jianbo and Soo Park, Hyun},
  booktitle={Proceedings of the IEEE conference on computer vision and pattern recognition},
  pages={1501--1510},
  year={2017}
}

@inproceedings{gazemissteps,
  title={Gazing into missteps: Leveraging eye-gaze for unsupervised mistake detection in egocentric videos of skilled human activities},
  author={Mazzamuto, Michele and Furnari, Antonino and Sato, Yoichi and Farinella, Giovanni Maria},
  booktitle={Proceedings of the Computer Vision and Pattern Recognition Conference},
  pages={8310--8320},
  year={2025}
}

@inproceedings{ListenToLookIntoFuture,
  title={Listen to look into the future: Audio-visual egocentric gaze anticipation},
  author={Lai, Bolin and Ryan, Fiona and Jia, Wenqi and Liu, Miao and Rehg, James M},
  booktitle={European Conference on Computer Vision},
  pages={192--210},
  year={2024},
  organization={Springer}
}

@inproceedings{PredictEgoGaze,
  title={Learning to predict gaze in egocentric video},
  author={Li, Yin and Fathi, Alireza and Rehg, James M},
  booktitle={Proceedings of the IEEE international conference on computer vision},
  pages={3216--3223},
  year={2013}
}

@inproceedings{GIMO,
  title={Gimo: Gaze-informed human motion prediction in context},
  author={Zheng, Yang and Yang, Yanchao and Mo, Kaichun and Li, Jiaman and Yu, Tao and Liu, Yebin and Liu, C Karen and Guibas, Leonidas J},
  booktitle={European Conference on Computer Vision},
  pages={676--694},
  year={2022},
  organization={Springer}
}

@inproceedings{GazeActionPredict,
  title={Gaze-Guided Graph Neural Network for Action Anticipation Conditioned on Intention},
  author={{\"O}zdel, S{\"u}leyman and Rong, Yao and Albaba, Berat Mert and Kuo, Yen-Ling and Wang, Xi and Kasneci, Enkelejda},
  booktitle={Proceedings of the 2024 Symposium on Eye Tracking Research and Applications},
  pages={1--9},
  year={2024}
}

@INPROCEEDINGS{GazeMotion,
  author={Anvari, Taravat and Lappe, Markus and de Lussanet, Marc H E},
  booktitle={2025 IEEE Conference on Virtual Reality and 3D User Interfaces Abstracts and Workshops (VRW)}, 
  title={Where Does Gaze Lead? Integrating Gaze and Motion for Enhanced 3D Pose Estimation}, 
  year={2025},
  volume={},
  number={},
  pages={76-83},
  doi={10.1109/VRW66409.2025.00025}}

@article{EyeofTransformer,
  title={In the eye of transformer: Global-local correlation for egocentric gaze estimation},
  author={Lai, Bolin and Liu, Miao and Ryan, Fiona and Rehg, James M},
  journal={arXiv preprint arXiv:2208.04464},
  year={2022}
}

@article{EyeBeholder,
  title={In the eye of the beholder: Gaze and actions in first person video},
  author={Li, Yin and Liu, Miao and Rehg, James M},
  journal={IEEE transactions on pattern analysis and machine intelligence},
  volume={45},
  number={6},
  pages={6731--6747},
  year={2021},
  publisher={IEEE}
}

@article{MCN,
  title={Mutual context network for jointly estimating egocentric gaze and action},
  author={Huang, Yifei and Cai, Minjie and Li, Zhenqiang and Lu, Feng and Sato, Yoichi},
  journal={IEEE Transactions on Image Processing},
  volume={29},
  pages={7795--7806},
  year={2020},
  publisher={IEEE}
}

@inproceedings{gaze_att_i3d,
  title={Integrating human gaze into attention for egocentric activity recognition},
  author={Min, Kyle and Corso, Jason J},
  booktitle={Proceedings of the IEEE/CVF Winter Conference on Applications of Computer Vision},
  pages={1069--1078},
  year={2021}
}

@inproceedings{GazeFromTask,
  title={Predicting gaze in egocentric video by learning task-dependent attention transition},
  author={Huang, Yifei and Cai, Minjie and Li, Zhenqiang and Sato, Yoichi},
  booktitle={Proceedings of the European conference on computer vision (ECCV)},
  pages={754--769},
  year={2018}
}

@inproceedings{YoudoIlearn,
  title={You-Do, I-Learn: Discovering Task Relevant Objects and their Modes of Interaction from Multi-User Egocentric Video.},
  author={Damen, Dima and Leelasawassuk, Teesid and Haines, Osian and Calway, Andrew and Mayol-Cuevas, Walterio W},
  booktitle={BMVC},
  volume={2},
  pages={3},
  year={2014}
}

@inproceedings{privaceye,
  title={Privaceye: privacy-preserving head-mounted eye tracking using egocentric scene image and eye movement features},
  author={Steil, Julian and Koelle, Marion and Heuten, Wilko and Boll, Susanne and Bulling, Andreas},
  booktitle={Proceedings of the 11th ACM symposium on eye tracking research \& applications},
  pages={1--10},
  year={2019}
}

@article{AttendedObjects,
  title={Learning to detect attended objects in cultural sites with gaze signals and weak object supervision},
  author={Mazzamuto*, Michele and Ragusa*, Francesco and Furnari*, Antonino and Farinella*, Giovanni Maria},
  journal={ACM Journal on Computing and Cultural Heritage},
  volume={17},
  number={3},
  pages={1--21},
  year={2024},
  publisher={ACM New York, NY}
}

@article{gazegpt,
  title={Gazegpt: Augmenting human capabilities using gaze-contingent contextual ai for smart eyewear},
  author={Konrad, Robert and Padmanaban, Nitish and Buckmaster, J Gabriel and Boyle, Kevin C and Wetzstein, Gordon},
  journal={arXiv preprint arXiv:2401.17217},
  year={2024}
}

@article{egotrigger,
  title={EgoTrigger: Toward Audio-Driven Image Capture for Human Memory Enhancement in All-Day Energy-Efficient Smart Glasses},
  author={Paruchuri, Akshay and Hersek, Sinan and Aggarwal, Lavisha and Yang, Qiao and Liu, Xin and Kulshrestha, Achin and Colaco, Andrea and Fuchs, Henry and Chatterjee, Ishan},
  journal={arXiv preprint arXiv:2508.01915},
  year={2025}
}

@inproceedings{listentolook,
  title={Listen to look: Action recognition by previewing audio},
  author={Gao, Ruohan and Oh, Tae-Hyun and Grauman, Kristen and Torresani, Lorenzo},
  booktitle={Proceedings of the IEEE/CVF conference on computer vision and pattern recognition},
  pages={10457--10467},
  year={2020}
}

@article{egodistill,
  title={Egodistill: Egocentric head motion distillation for efficient video understanding},
  author={Tan, Shuhan and Nagarajan, Tushar and Grauman, Kristen},
  journal={Advances in Neural Information Processing Systems},
  volume={36},
  pages={33485--33498},
  year={2023}
}

@inproceedings{chat2map,
  title={Chat2map: Efficient scene mapping from multi-ego conversations},
  author={Majumder, Sagnik and Jiang, Hao and Moulon, Pierre and Henderson, Ethan and Calamia, Paul and Grauman, Kristen and Ithapu, Vamsi Krishna},
  booktitle={Proceedings of the IEEE/CVF Conference on Computer Vision and Pattern Recognition},
  pages={10554--10564},
  year={2023}
}

@article{smartapm,
  title={SmartAPM framework for adaptive power management in wearable devices using deep reinforcement learning},
  author={Sunder, R and Lilhore, Umesh Kumar and Rai, Anjani Kumar and Ghith, Ehab and Tlija, Mehdi and Simaiya, Sarita and Majeed, Afraz Hussain},
  journal={Scientific Reports},
  volume={15},
  number={1},
  pages={6911},
  year={2025},
  publisher={Nature Publishing Group UK London}
}

@ARTICLE{distributedcomputation,
  author={Wang, Chaowei and Wang, Ziye and Guan, Weiwei and Wang, Wenjie and Xu, Lexi and Li, Lihua and Huang, Sai and Wang, Weidong},
  journal={IEEE Transactions on Consumer Electronics}, 
  title={Trustworthy Health Monitoring Based on Distributed Wearable Electronics With Edge Intelligence}, 
  year={2024},
  volume={70},
  number={1},
  pages={2333-2341},
  doi={10.1109/TCE.2024.3358803}}

@inproceedings{scsampler,
  title={Scsampler: Sampling salient clips from video for efficient action recognition},
  author={Korbar, Bruno and Tran, Du and Torresani, Lorenzo},
  booktitle={Proceedings of the IEEE/CVF International Conference on Computer Vision},
  pages={6232--6242},
  year={2019}
}

@inproceedings{flexibleframe,
  title={Flexible Frame Selection for Efficient Video Reasoning},
  author={Buch, Shyamal and Nagrani, Arsha and Arnab, Anurag and Schmid, Cordelia},
  booktitle={Proceedings of the Computer Vision and Pattern Recognition Conference},
  pages={29071--29082},
  year={2025}
}

@inproceedings{X3d,
  title={X3d: Expanding architectures for efficient video recognition},
  author={Feichtenhofer, Christoph},
  booktitle={Proceedings of the IEEE/CVF conference on computer vision and pattern recognition},
  pages={203--213},
  year={2020}
}

@inproceedings{lightASDNet,
  title={A light weight model for active speaker detection},
  author={Liao, Junhua and Duan, Haihan and Feng, Kanghui and Zhao, Wanbing and Yang, Yanbing and Chen, Liangyin},
  booktitle={Proceedings of the IEEE/CVF conference on computer vision and pattern recognition},
  pages={22932--22941},
  year={2023}
}

@article{psy_gazesport,
  title={Gaze Control and Motor Performance in Motor Expertise Studies: Focused Review of Field Application Research on Perceptual Skill Training.},
  author={Lee, Seungmin and An, Jongseong},
  journal={International Journal of Applied Sports Sciences},
  volume={35},
  number={1},
  year={2023}
}

@article{psy_surgerytraining,
  title={Gaze behavior is related to objective technical skills assessment during virtual reality simulator-based surgical training: a proof of concept},
  author={Galuret, Soline and Vall{\'e}e, Nicolas and Tronchot, Alexandre and Thomazeau, Herve and Jannin, Pierre and Huaulm{\'e}, Arnaud},
  journal={International Journal of Computer Assisted Radiology and Surgery},
  volume={18},
  number={9},
  pages={1697--1705},
  year={2023},
  publisher={Springer}
}

@article{psy_medicaleducation,
  title={A review of eye tracking for understanding and improving diagnostic interpretation},
  author={Bruny{\'e}, Tad T and Drew, Trafton and Weaver, Donald L and Elmore, Joann G},
  journal={Cognitive research: principles and implications},
  volume={4},
  number={1},
  pages={7},
  year={2019},
  publisher={Springer}
}

@inproceedings{prosandcons,
  title={The pros and cons: Rank-aware temporal attention for skill determination in long videos},
  author={Doughty, Hazel and Mayol-Cuevas, Walterio and Damen, Dima},
  booktitle={Proceedings of the IEEE/CVF conference on computer vision and pattern recognition},
  pages={7862--7871},
  year={2019}
}

@article{psy_robotsurgery,
  title={See like an expert: Gaze-augmented training enhances skill acquisition in a virtual reality robotic suturing task},
  author={Melnyk, Rachel and Campbell, Timothy and Holler, Tyler and Cameron, Katherine and Saba, Patrick and Witthaus, Michael W and Joseph, Jean and Ghazi, Ahmed},
  journal={Journal of Endourology},
  volume={35},
  number={3},
  pages={376--382},
  year={2021},
  publisher={Mary Ann Liebert, Inc., publishers 140 Huguenot Street, 3rd Floor New~…}
}

@article{psy_LookAheadFixations,
  title={Look-ahead fixations during visuomotor behavior: Evidence from assembling a camping tent},
  author={Sullivan, Brian and Ludwig, Casimir JH and Damen, Dima and Mayol-Cuevas, Walterio and Gilchrist, Iain D},
  journal={Journal of vision},
  volume={21},
  number={3},
  pages={13--13},
  year={2021},
  publisher={The Association for Research in Vision and Ophthalmology}
}

@ARTICLE{psy_GazeHotspot,
author={Longfei CHEN and Yuichi NAKAMURA and Kazuaki KONDO and Dima DAMEN and Walterio MAYOL-CUEVAS},
journal={IEICE TRANSACTIONS on Information},
title={Integration of Experts' and Beginners' Machine Operation Experiences to Obtain a Detailed Task Model},
year={2021},
volume={E104-D},
number={1},
pages={152-161},
doi={10.1587/transinf.2019EDP7180},
ISSN={1745-1361},
month={January},}

@article{psy_dailyroles,
  title={The roles of vision and eye movements in the control of activities of daily living},
  author={Land, Michael and Mennie, Neil and Rusted, Jennifer},
  journal={Perception},
  volume={28},
  number={11},
  pages={1311--1328},
  year={1999},
  publisher={SAGE Publications Sage UK: London, England}
}

@article{psy_esports,
  title={Difference in gaze control ability between low and high skill players of a real-time strategy game in esports},
  author={Jeong, Inhyeok and Nakagawa, Kento and Osu, Rieko and Kanosue, Kazuyuki},
  journal={PloS one},
  volume={17},
  number={3},
  pages={e0265526},
  year={2022},
  publisher={Public Library of Science San Francisco, CA USA}
}

@article{psy_decisionmaking,
  title={Using eye tracking to trace a cognitive process: Gaze behaviour during decision making in a natural environment},
  author={Gidl{\"o}f, Kerstin and Wallin, Annika and Dewhurst, Richard and Holmqvist, Kenneth},
  journal={Journal of eye movement research},
  volume={6},
  number={1},
  year={2013}
}

@inproceedings{score_skill,
  title={Uncertainty-aware score distribution learning for action quality assessment},
  author={Tang, Yansong and Ni, Zanlin and Zhou, Jiahuan and Zhang, Danyang and Lu, Jiwen and Wu, Ying and Zhou, Jie},
  booktitle={Proceedings of the IEEE/CVF conference on computer vision and pattern recognition},
  pages={9839--9848},
  year={2020}
}

@article{psy_GazeAndGoal,
  title={Control of gaze in natural environments: effects of rewards and costs, uncertainty and memory in target selection},
  author={Hayhoe, Mary M and Matthis, Jonathan Samir},
  journal={Interface focus},
  volume={8},
  number={4},
  pages={20180009},
  year={2018},
  publisher={The Royal Society}
}

@article{whentosay,
  title={What to say and when to say it: Live fitness coaching as a testbed for situated interaction},
  author={Panchal, Sunny and Bhattacharyya, Apratim and Berger, Guillaume and Mercier, Antoine and B{\"o}hm, Cornelius and Dietrichkeit, Florian and Pourreza, Reza and Li, Xuanlin and Madan, Pulkit and Lee, Mingu and others},
  journal={Advances in Neural Information Processing Systems},
  volume={37},
  pages={75853--75882},
  year={2024}
}

@article{exact,
  title={ExAct: A Video-Language Benchmark for Expert Action Analysis},
  author={Yi, Han and Pan, Yulu and He, Feihong and Liu, Xinyu and Zhang, Benjamin and Oguntola, Oluwatumininu and Bertasius, Gedas},
  journal={arXiv preprint arXiv:2506.06277},
  year={2025}
}

@inproceedings{ProgressAssessment,
author = {Drey, Tobias and Jansen, Pascal and Fischbach, Fabian and Frommel, Julian and Rukzio, Enrico},
title = {Towards Progress Assessment for Adaptive Hints in Educational Virtual Reality Games},
year = {2020},
isbn = {9781450368193},
publisher = {Association for Computing Machinery},
address = {New York, NY, USA},
url = {https://doi.org/10.1145/3334480.3382789},
doi = {10.1145/3334480.3382789},
booktitle = {Extended Abstracts of the 2020 CHI Conference on Human Factors in Computing Systems},
pages = {1–9},
numpages = {9},
location = {Honolulu, HI, USA},
series = {CHI EA '20}
}

@inproceedings{eyepiano,
  title={EyePiano: leveraging gaze for reflective piano learning},
  author={Karolus, Jakob and Sylupp, Johannes and Schmidt, Albrecht and Wo{\'z}niak, Pawe{\l} W},
  booktitle={Proceedings of the 2023 ACM Designing Interactive Systems Conference},
  pages={1209--1223},
  year={2023}
}

@InProceedings{BASKET_CVPR25,
  author = {Pan, Yulu and Zhang, Ce and Bertasius, Gedas},
  title = {BASKET: A Large-Scale Video Dataset for Fine-Grained Skill Estimation},
  booktitle = {Proceedings of the IEEE/CVF Conference on Computer Vision and Pattern Recognition (CVPR)},
  month = {June},
  year = {2025}
}

@article{videodiff,
  title={Video Action Differencing},
  author={Burgess, James and Wang, Xiaohan and Zhang, Yuhui and Rau, Anita and Lozano, Alejandro and Dunlap, Lisa and Darrell, Trevor and Yeung-Levy, Serena},
  journal={arXiv preprint arXiv:2503.07860},
  year={2025}
}

@inproceedings{expertaf,
  title={ExpertAF: Expert actionable feedback from video},
  author={Ashutosh, Kumar and Nagarajan, Tushar and Pavlakos, Georgios and Kitani, Kris and Grauman, Kristen},
  booktitle={Proceedings of the Computer Vision and Pattern Recognition Conference},
  pages={13582--13594},
  year={2025}
}

@inproceedings{multitask_assessment,
  title={What and how well you performed? a multitask learning approach to action quality assessment},
  author={Parmar, Paritosh and Morris, Brendan Tran},
  booktitle={Proceedings of the IEEE/CVF conference on computer vision and pattern recognition},
  pages={304--313},
  year={2019}
}

@inproceedings{logo,
  title={Logo: A long-form video dataset for group action quality assessment},
  author={Zhang, Shiyi and Dai, Wenxun and Wang, Sujia and Shen, Xiangwei and Lu, Jiwen and Zhou, Jie and Tang, Yansong},
  booktitle={Proceedings of the IEEE/CVF conference on computer vision and pattern recognition},
  pages={2405--2414},
  year={2023}
}

@inproceedings{expert_novice_soccer,
  title={Classification of expert-novice level using eye tracking and motion data via conditional multimodal variational autoencoder},
  author={Akamatsu, Yusuke and Maeda, Keisuke and Ogawa, Takahiro and Haseyama, Miki},
  booktitle={ICASSP 2021-2021 IEEE International Conference on Acoustics, Speech and Signal Processing (ICASSP)},
  pages={1360--1364},
  year={2021},
  organization={IEEE}
}

@article{IMU_Skill,
  title={Generalized and efficient skill assessment from IMU data with applications in gymnastics and medical training},
  author={Khan, Aftab and Mellor, Sebastian and King, Rachel and Janko, Balazs and Harwin, William and Sherratt, R Simon and Craddock, Ian and Pl{\"o}tz, Thomas},
  journal={ACM Transactions on Computing for Healthcare},
  volume={2},
  number={1},
  pages={1--21},
  year={2020},
  publisher={ACM New York, NY, USA}
}

@article{music_eyehand,
  title={Review on eye-hand span in sight-reading of music},
  author={Perra, Joris and Poulin-Charronnat, B{\'e}n{\'e}dicte and Baccino, Thierry and Drai-Zerbib, V{\'e}ronique},
  journal={Journal of eye movement research},
  volume={14},
  number={4},
  pages={10--16910},
  year={2021}
}

@misc{projectaria_glasses_manual,
  author       = {{Meta Platforms, Inc.}},
  title        = {Project Aria Glasses User Manual},
  howpublished = {\url{https://facebookresearch.github.io/projectaria_tools/docs/ARK/glasses_manual/glasses_user_manual}},
  note         = {Accessed: 2025-10-06},
  year         = {2025}
}

@article{camera_power,
  title={Low power environmental image sensors for remote photogrammetry},
  author={Balde, Alpha Yaya and Bergeret, Emmanuel and Cajal, Denis and Toumazet, Jean-Pierre},
  journal={Sensors},
  volume={22},
  number={19},
  pages={7617},
  year={2022},
  publisher={MDPI}
}

@article{camera_power2,
  title={HyperCam: Low-Power Onboard Computer Vision for IoT Cameras},
  author={Lee, Chae Young and Fite, Maxwell and Rao, Tejus and Achour, Sara and Kapetanovic, Zerina and others},
  journal={arXiv preprint arXiv:2501.10547},
  year={2025}
}

@inproceedings{deit,
  title={Training data-efficient image transformers \& distillation through attention},
  author={Touvron, Hugo and Cord, Matthieu and Douze, Matthijs and Massa, Francisco and Sablayrolles, Alexandre and J{\'e}gou, Herv{\'e}},
  booktitle={International conference on machine learning},
  pages={10347--10357},
  year={2021},
  organization={PMLR}
}

@misc{fitnets,
      title={FitNets: Hints for Thin Deep Nets}, 
      author={Adriana Romero and Nicolas Ballas and Samira Ebrahimi Kahou and Antoine Chassang and Carlo Gatta and Yoshua Bengio},
      year={2015},
      eprint={1412.6550},
      archivePrefix={arXiv},
      primaryClass={cs.LG},
      url={https://arxiv.org/abs/1412.6550}, 
}

@inproceedings{distilling,
  title={Distilling object detectors with fine-grained feature imitation},
  author={Wang, Tao and Yuan, Li and Zhang, Xiaopeng and Feng, Jiashi},
  booktitle={Proceedings of the IEEE/CVF conference on computer vision and pattern recognition},
  pages={4933--4942},
  year={2019}
}

@article{vid2coach,
  title={Vid2Coach: Transforming How-To Videos into Task Assistants},
  author={Huh, Mina and Xue, Zihui and Das, Ujjaini and Ashutosh, Kumar and Grauman, Kristen and Pavel, Amy},
  journal={arXiv preprint arXiv:2506.00717},
  year={2025}
}

@article{egoppg,
  title={egoppg: Heart rate estimation from eye-tracking cameras in egocentric systems to benefit downstream vision tasks},
  author={Braun, Bj{\"o}rn and Armani, Rayan and Meier, Manuel and Moebus, Max and Holz, Christian},
  journal={arXiv preprint arXiv:2502.20879},
  year={2025}
}

@article{badminton,
  title={Multisensebadminton: Wearable sensor--based biomechanical dataset for evaluation of badminton performance},
  author={Seong, Minwoo and Kim, Gwangbin and Yeo, Dohyeon and Kang, Yumin and Yang, Heesan and DelPreto, Joseph and Matusik, Wojciech and Rus, Daniela and Kim, SeungJun},
  journal={Scientific Data},
  volume={11},
  number={1},
  pages={343},
  year={2024},
  publisher={Nature Publishing Group UK London}
}

@Article{psy_eyespan,
    AUTHOR = {Cara, Michel A.},
    TITLE = {The Effect of Practice and Musical Structure on Pianists’ Eye-Hand Span and Visual Monitoring},
    JOURNAL = {Journal of Eye Movement Research},
    VOLUME = {16},
    YEAR = {2023},
    NUMBER = {2},
    PAGES = {1--18},
    URL = {https://www.mdpi.com/1995-8692/16/2/11},
    ISSN = {1995-8692},
    DOI = {10.16910/jemr.16.2.5}
}

@inproceedings{Timesformer,
  title={Is space-time attention all you need for video understanding?},
  author={Bertasius, Gedas and Wang, Heng and Torresani, Lorenzo},
  booktitle={Icml},
  volume={2},
  number={3},
  pages={4},
  year={2021}
}

@misc{skillformer,
      title={SkillFormer: Unified Multi-View Video Understanding for Proficiency Estimation}, 
      author={Edoardo Bianchi and Antonio Liotta},
      year={2025},
      eprint={2505.08665},
      archivePrefix={arXiv},
      primaryClass={cs.CV},
      url={https://arxiv.org/abs/2505.08665}, 
}

@inproceedings{e2gomotion,
  title={E2 (go) motion: Motion augmented event stream for egocentric action recognition},
  author={Plizzari, Chiara and Planamente, Mirco and Goletto, Gabriele and Cannici, Marco and Gusso, Emanuele and Matteucci, Matteo and Caputo, Barbara},
  booktitle={Proceedings of the IEEE/CVF conference on computer vision and pattern recognition},
  pages={19935--19947},
  year={2022}
}

@article{egovlpv2,
  title={EgoVLPv2: Egocentric Video-Language Pre-training with Fusion in the Backbone},
  author={Pramanick, Shraman and Song, Yale and Nag, Sayan and Lin, Kevin Qinghong and Shah, Hardik and Shou, Mike Zheng and Chellappa, Rama and Zhang, Pengchuan},
  journal={arXiv preprint arXiv:2307.05463},
  year={2023}
}

@article{dinov2,
  title={Dinov2: Learning robust visual features without supervision},
  author={Oquab, Maxime and Darcet, Timoth{\'e}e and Moutakanni, Th{\'e}o and Vo, Huy and Szafraniec, Marc and Khalidov, Vasil and Fernandez, Pierre and Haziza, Daniel and Massa, Francisco and El-Nouby, Alaaeldin and others},
  journal={arXiv preprint arXiv:2304.07193},
  year={2023}
}

@article{power_alpha,
title = {Trends in AI inference energy consumption: Beyond the performance-vs-parameter laws of deep learning},
journal = {Sustainable Computing: Informatics and Systems},
volume = {38},
pages = {100857},
year = {2023},
issn = {2210-5379},
doi = {https://doi.org/10.1016/j.suscom.2023.100857},
url = {https://www.sciencedirect.com/science/article/pii/S2210537923000124},
author = {Radosvet Desislavov and Fernando Martínez-Plumed and José Hernández-Orallo}
}

@INPROCEEDINGS{power_beta,
  author={Horowitz, Mark},
  booktitle={2014 IEEE International Solid-State Circuits Conference Digest of Technical Papers (ISSCC)}, 
  title={1.1 Computing's energy problem (and what we can do about it)}, 
  year={2014},
  volume={},
  number={},
  pages={10-14},
  doi={10.1109/ISSCC.2014.6757323}}

@article{psy_soccer_intro,
  author       = {},
  title        = {Visual strategies of young soccer players during a passing test – A pilot study},
  journal      = {Journal of Eye Movement Research},
  year         = {2022},
  month        = {Feb},
  volume       = {15},
  number       = {1},
  pages        = {},
  doi          = {10.16910/jemr.15.1.3},
  url          = {https://bop.unibe.ch/JEMR/article/view/8148}
}

@inproceedings{i3d,
  title={Quo vadis, action recognition? a new model and the kinetics dataset},
  author={Carreira, Joao and Zisserman, Andrew},
  booktitle={proceedings of the IEEE Conference on Computer Vision and Pattern Recognition},
  pages={6299--6308},
  year={2017}
}

@article{lora,
  title={Lora: Low-rank adaptation of large language models.},
  author={Hu, Edward J and Shen, Yelong and Wallis, Phillip and Allen-Zhu, Zeyuan and Li, Yuanzhi and Wang, Shean and Wang, Lu and Chen, Weizhu and others},
  journal={ICLR},
  volume={1},
  number={2},
  pages={3},
  year={2022}
}

@article{power_numbers,
  title={Advancements in Context Recognition for Edge Devices and Smart Eyewear: Sensors and Applications},
  author={Palermo, Francesca and Casciano, Luca and Demagh, Lokmane and Teliti, Aurelio and Antonello, Niccol{\`o} and Gervasoni, Giacomo and Shalby, Hazem Hesham Yousef and Paracchini, Marco Brando and Mentasti, Simone and Quan, Hao and others},
  journal={IEEE Access},
  year={2025},
  publisher={IEEE}
}

@article{electrasight,
  title={ElectraSight: Fully Onboard Eye Tracking for Smart Glasses With Hybrid EOG (hEOG)},
  author={Sch{\"a}rer, Nicolas and Villani, Federico and Melatur, Aishwarya and Peter, Steven and Polonelli, Tommaso and Magno, Michele},
  journal={IEEE Internet of Things Journal},
  year={2025},
  publisher={IEEE}
}

@article{evostruggle,
  title={EvoStruggle: A Dataset Capturing the Evolution of Struggle across Activities and Skill Levels},
  author={Feng, Shijia and Wray, Michael and Mayol-Cuevas, Walterio},
  journal={arXiv preprint arXiv:2510.01362},
  year={2025}
}

@article{skill_level,
  author    = {Isabel Funke and Sören Torge Mees and Jürgen Weitz and Stefanie Speidel},
  title     = {Video-based surgical skill assessment using 3D convolutional neural networks},
  journal   = {International Journal of Computer Assisted Radiology and Surgery},
  year      = {2019},
  volume    = {14},
  number    = {7},
  pages     = {1217--1225},
  doi       = {10.1007/s11548-019-01995-1},
  url       = {https://doi.org/10.1007/s11548-019-01995-1},
  issn      = {1861-6429}
}

@inproceedings{piano_skill_level,
  title={Piano skills assessment},
  author={Parmar, Paritosh and Reddy, Jaiden and Morris, Brendan},
  booktitle={2021 IEEE 23rd international workshop on multimedia signal processing (MMSP)},
  pages={1--5},
  year={2021},
  organization={IEEE}
}

@inproceedings{holoassist,
  title={Holoassist: an egocentric human interaction dataset for interactive ai assistants in the real world},
  author={Wang, Xin and Kwon, Taein and Rad, Mahdi and Pan, Bowen and Chakraborty, Ishani and Andrist, Sean and Bohus, Dan and Feniello, Ashley and Tekin, Bugra and Frujeri, Felipe Vieira and others},
  booktitle={Proceedings of the IEEE/CVF International Conference on Computer Vision},
  pages={20270--20281},
  year={2023}
}

@article{egoblind,
  title={EgoBlind: Towards Egocentric Visual Assistance for the Blind People},
  author={Xiao, Junbin and Huang, Nanxin and Qiu, Hao and Tao, Zhulin and Yang, Xun and Hong, Richang and Wang, Meng and Yao, Angela},
  journal={arXiv preprint arXiv:2503.08221},
  year={2025}
}

@InProceedings{egotextvqa,
    author    = {Zhou, Sheng and Xiao, Junbin and Li, Qingyun and Li, Yicong and Yang, Xun and Guo, Dan and Wang, Meng and Chua, Tat-Seng and Yao, Angela},
    title     = {EgoTextVQA: Towards Egocentric Scene-Text Aware Video Question Answering},
    booktitle = {Proceedings of the IEEE/CVF Conference on Computer Vision and Pattern Recognition (CVPR)},
    month     = {June},
    year      = {2025},
    pages     = {3363-3373}
}
}

\clearpage
\appendix
\setcounter{page}{1}
\maketitlesupplementary

\section{Supplementary video}

We provide a supplementary video that shows an overview of the paper. It also shows qualitative video examples with ego video and gaze patterns.

\section{Ablations}
\textbf{SkillSight-T.} In Sec.~\ref{sec:teacher}, we present the three components of SkillSight-T: a visual encoder with gaze attention, a cropped image encoder, and a trajectory encoder. We perform an ablation study on these designs using EgoExo4D~\cite{egoexo4d}. As shown in Table~\ref{tab:ablation}, each component contributes a clear gain in accuracy, indicating that these designs are essential for capturing the interaction between ego visual input and gaze in skill assessment.

\textbf{SkillSight-S.} In Sec.~\ref{sec:distill}, we introduce the distillation strategy used for training SkillSight-S. We evaluate the influence of each loss through an ablation study, with results summarized in Table~\ref{tab:distill_ablation}. The results demonstrate that both the distillation loss and the action recognition loss contribute positively to the performance of SkillSight-S. This highlights the importance of linking gaze patterns with specific actions for effective skill assessment, and video cues for skill assessment can be effectively embedded into the gaze signal.

\JWCam{\textbf{Input modalities.} While SkillSight-S leverages both gaze and head motion as inputs, we further examine the contribution of each modality by separating gaze direction and head rotation. Using the same distillation training, SkillSight-S achieves $41.4$ with gaze-only, $41.8$ with head-motion-only, and $44.4$ with both head-motion and gaze, demonstrating that both modalities are necessary.}

\begin{table}[ht]
\centering
\small
\begin{tabular}{lcc c}
\toprule
TimeSformer & Crop Encoder & Traj. Encoder & Acc (\%) \\
\midrule
{\xmark} & {\xmark} & {\cmark} & 37.0 \\
{\xmark} & {\cmark} & {\xmark} & 40.6 \\
{\xmark} & {\cmark} & {\cmark} & 44.9 \\ % example placeholder
w/o gaze att. & {\xmark} & {\xmark} & 45.5 \\
w/o gaze att. & {\xmark} & {\cmark} & 46.4 \\
w/o gaze att. & {\cmark} & {\xmark} & 47.6\\
w/o gaze att. & {\cmark} & {\cmark} & 47.9\\
w gaze att. & {\xmark} & {\xmark} & 47.2 \\
w gaze att. & {\xmark} & {\cmark} & 47.5 \\
w gaze att. & {\cmark} & {\xmark} & 48.6 \\ % example 
w gaze att. & {\cmark} & {\cmark} & \textbf{50.1} \\
\bottomrule
\end{tabular}
\caption{\textbf{Ablation study of SkillSight-T.} We conduct ablation study of the three components in SkillSight-T. A check indicates inclusion.}
\label{tab:ablation}
\end{table}

\begin{table}[ht]
\centering
\small
\begin{tabular}{lc}
\toprule
Method & Acc (\%) \\
\midrule
Gaze-Only & 37.0 \\
SkillSight-S & \textbf{44.4} \\
w/o distillation & 40.0 \\
w/o action recognition & 40.7 \\
\bottomrule
\end{tabular}
\caption{\textbf{Ablation study of training SkillSight-S.} We compare the performance of SkillSight-S when training under different loss configurations.}
\label{tab:distill_ablation}
\end{table}

\section{Gaze normalization process}
In Sec.~\ref{sec:problem}, we describe the gaze modalities derived from the three-dimensional gaze vector of each eye. We also provide details on how each modality is normalized to remove bias in the recordings.

\begin{itemize}

\item \textbf{3D fixation points.}
For each frame, we calculate the intersection of the left and right gaze rays in the world coordinate. Centered by the segment mean and rotated horizontally so the first direction gaze point has y equal to zero.

\item \textbf{3D gaze direction.}
For each frame, this is a unit vector representing the gaze direction expressed in the cpf coordinate as defined by the subject's perspective~\cite{projectaria_glasses_manual}.

\item \textbf{2D gaze point projection.}
We project the 3D gaze to the 2D ego camera view, value in the range zero to one.

\item \textbf{Gaze depth.}
Distance between the head and the intersection point of the left and right gaze rays.

\item \textbf{Glass rotation.}
For the first frame, adjust the yaw to face forward while keeping pitch and roll. For other frames, compute relative rotation. Convert the rotation representation from pitch, yaw, roll to quaternion.

\item \textbf{Glass translation.}
Center the translation and define the first horizontal movement as the positive x direction.

\end{itemize}

\JWCam{A modality is included in an experiment if the corresponding dataset provides it. In EgoExo4D~\cite{egoexo4d}, the glasses translation is obtained using visual-inertial-odometry, which relies on a low-power SLAM camera. Regardless, a SLAM camera is not a strict requirement; trajectories can alternatively be inferred using IMU alone~\cite{power_numbers}.} We also evaluate a variant of SkillSight-S that uses only
head rotation and 3D gaze, and observe a performance of
44.4\% on EgoExo4D (a 0.4\% drop compared to when trajectory is enabled).

\section{Power consumption calculation details}
In Sec.~\ref{sec:expts}, we demonstrate the power efficiency analysis. Following the energy computation in \cite{egoexo4d}, the overall energy consumption rate consists of the following:

\begin{itemize}

\item \textbf{Compute operations (MACs).}
We estimate computational cost by using the PyTorch FLOP counter to measure the total FLOPs in a forward pass, and then convert this value to MACs using the approximation that one MAC equals two FLOPs.

\item \textbf{Memory transfer (bytes).}
We quantify GPU memory movement with the PyTorch memory profiler, which records all operations in the forward pass along with their memory usage. The total memory transfer is the sum of the memory costs of all logged operations.

\item \textbf{Sensor capture.}
For each sensing modality, we measure the period during which it is active by counting the number of samples that include that modality. We require at least one second of data, since energy consumption cannot be defined for a single instantaneous reading.

\end{itemize}

\begin{figure}[t] % [t] = top of column, can also use [h] or [b]
    \centering
    \includegraphics[width=\linewidth]{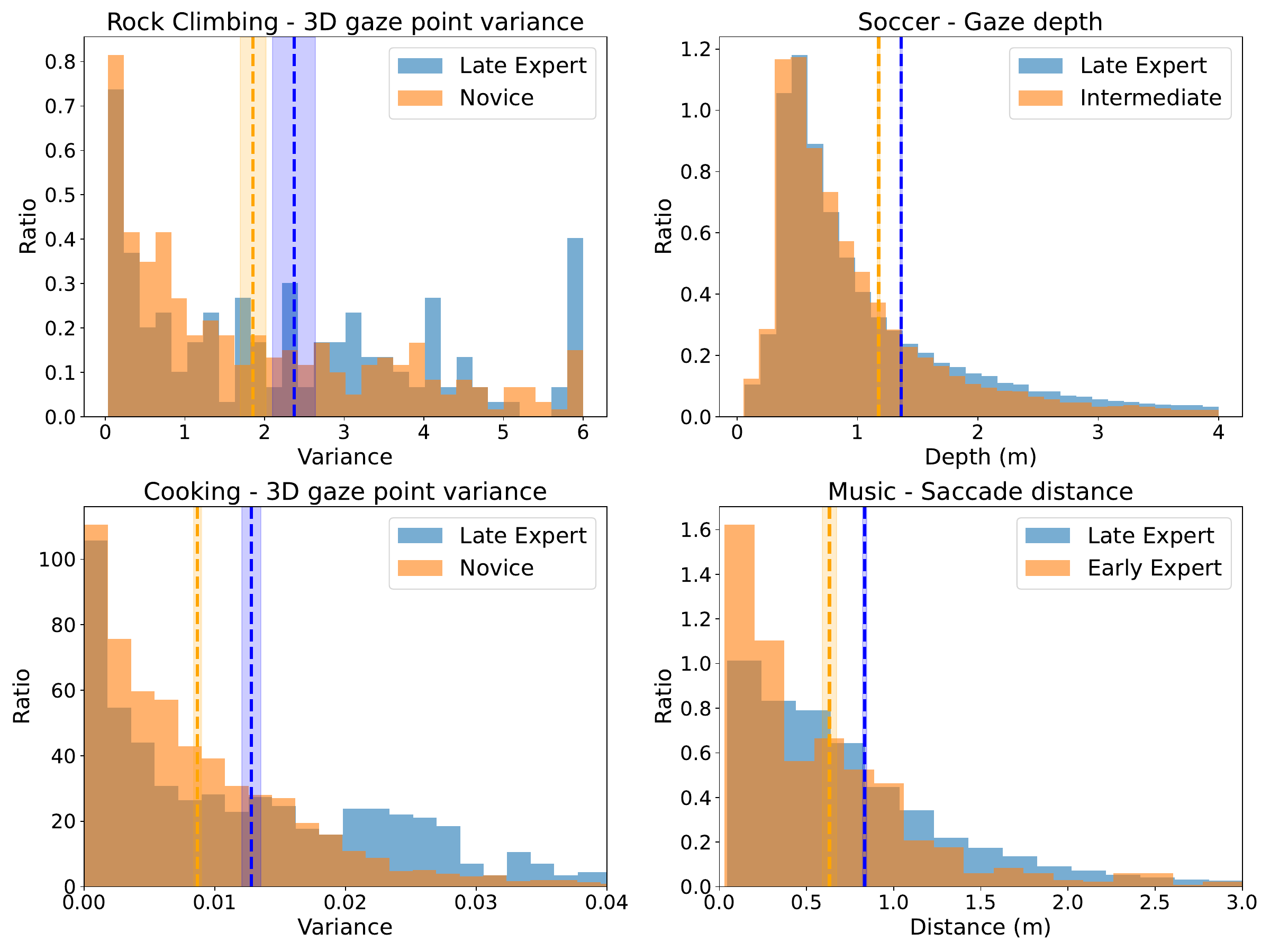}
    % Placeholder box
    % \fbox{\rule{0pt}{1.5in} \rule{0.9\columnwidth}{0pt}}
    % Caption
    \vspace{-0.7cm}
    \caption{\textbf{Distinct gaze pattern analysis.} We present more distinct gaze patterns that SkillSight-S reveals between subjects at different skill levels.}
    \label{fig:sup_analysis}
    \vspace{-0.3cm}
\end{figure}

\section{Behavior-level interpretation of gaze}
In Figure \ref{fig:analysis}, we present the distinct gaze patterns uncovered from the proficiency groups predicted by SkillSight-S. Figure \ref{fig:sup_analysis} further illustrates additional gaze behavior insights captured by SkillSight-S. In rock climbing, the variance of 3D gaze points is notably larger for the predicted late expert than for the novice, a result of frequent gaze shifts to gather information from the wall. In soccer, predicted late experts tend to fixate on farther depths, reflecting their attention to broader surroundings and potential targets. In cooking, predicted experienced chefs exhibit a more diverse 3D gaze points while monitoring food. Finally, in music, the model-predicted late experts switch more flexibly between the sheet music and the instrument, resulting in longer saccade distances. These findings enable deeper investigation of how gaze patterns vary across skill levels, allowing a more data-driven understanding of expertise.

\begin{figure}[t] % [t] = top of column, can also use [h] or [b]
    \centering
    \includegraphics[width=\linewidth]{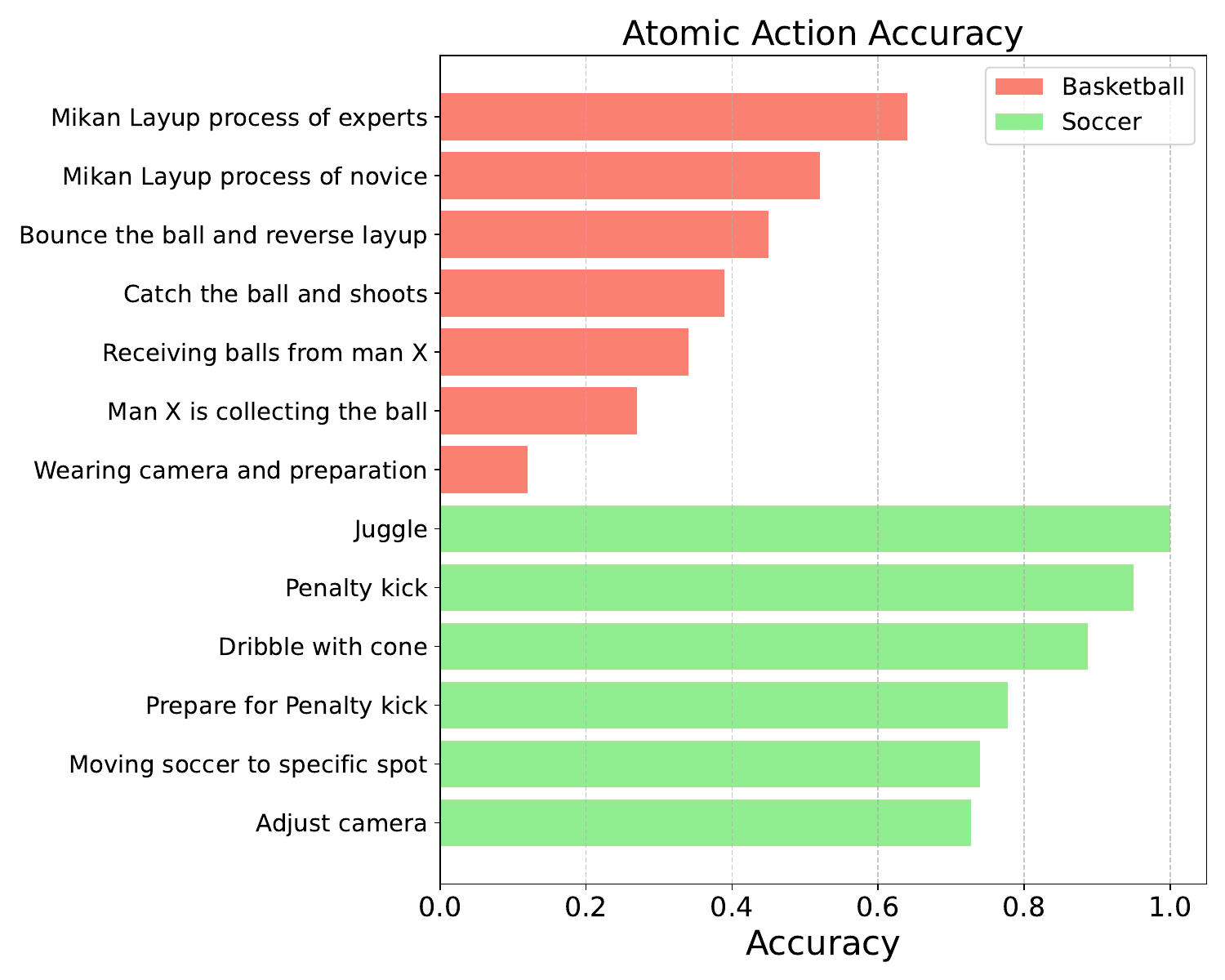}
    % Placeholder box
    % \fbox{\rule{0pt}{1.5in} \rule{0.9\columnwidth}{0pt}}
    % Caption
    \vspace{-0.7cm}
    \caption{\textbf{Atomic action analysis for SkillSight-S.} We present the performance of SkillSight-S across different atomic actions.}
    \label{fig:distill_analysis}
    \vspace{-0.3cm}
\end{figure}

\JWCam{
\section{Analysis of performance across actions}
In Figure \ref{fig:distill_analysis}, we analyze the performance of SkillSight-S across different actions by clustering Ego-Exo4D atomic action captions. We observe that gaze alone preserves most skill cues in perception-driven actions, such as basketball layups, soccer dribbling, and penalty kicks, where motion intent is strongly reflected in gaze. In contrast, performance degrades for actions requiring subtle motor execution, such as basketball shooting.
}

\end{document}